\documentclass[lettersize,journal]{IEEEtran}
\usepackage{amsmath,amsfonts}
\usepackage{algorithmic}
\usepackage{algorithm}
\usepackage{array}
\usepackage[caption=false,font=normalsize,labelfont=sf,textfont=sf]{subfig}
\usepackage{textcomp}
\usepackage{stfloats}
\usepackage{url}
\usepackage{verbatim}
\usepackage{graphicx}
\usepackage{cite}

\usepackage{color}
\usepackage{booktabs}
\usepackage{multirow} 
\usepackage{amsmath}
\usepackage{bbding}
\usepackage{makecell}
\usepackage{authblk}
\begin{document}

\title{Field-of-View IoU for Object Detection \\ in 360° Images}

\author[1]{Miao Cao}
\author[2]{Satoshi Ikehata}
\author[1]{Kiyoharu Aizawa}
\affil[1]{The University of Tokyo}
\affil[2]{National Institute of Informatics}



\maketitle

\begin{abstract}
360\textdegree~cameras have gained popularity over the last few years. In this paper, we propose two fundamental techniques---Field-of-View IoU (FoV-IoU) and 360Augmentation for object detection in 360\textdegree~images. Although most object detection neural networks designed for the perspective images are applicable to 360\textdegree~images in equirectangular projection (ERP) format, their performance deteriorates owing to the distortion in ERP images. Our method can be readily integrated with existing perspective object detectors and significantly improves the performance. The FoV-IoU computes the intersection-over-union of two Field-of-View bounding boxes in a spherical image which could be used for training, inference, and evaluation while 360Augmentation is a data augmentation technique specific to 360\textdegree~object detection task which randomly rotates a spherical image and solves the bias due to the sphere-to-plane projection. We conduct extensive experiments on the 360\textdegree~indoor dataset with different 
types of perspective object detectors and show the consistent effectiveness of our method.
\end{abstract}


\begin{figure*}[t]
\begin{center}
\includegraphics[width=0.9\linewidth]{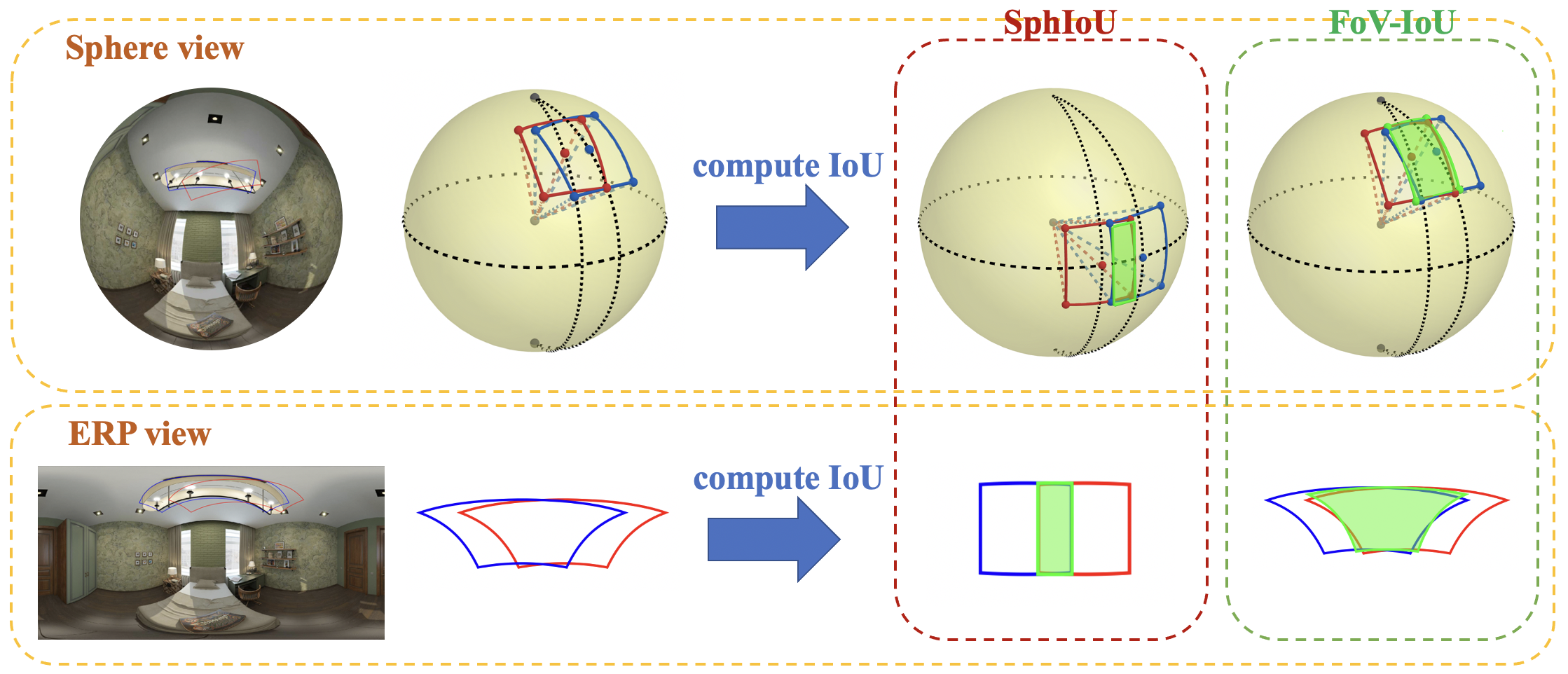}
\end{center}
   \caption{Illustration of Sph-IoU and FoV-IoU. The green bounding box is the intersection area calculated by Sph-IoU(left) and FoV-IoU(right) respectively. Because Sph-IoU makes the intersection area underestimated, we propose FoV-IoU to solve this problem and provide an accurate and efficient way to calculate the irregular intersection area.}
   \label{fig:sph_fov}
\end{figure*}

\section{Introduction}
360\textdegree~images have been used in many applications, such as virtual reality, autonomous driving and security monitoring. With the increase in 360\textdegree~image data, the demand for 360\textdegree~image recognition tasks, especially object detection tasks, also increases. 

One of the straightforward approaches for 360\textdegree~object detection is to project 360\textdegree~information to several perspective images and then apply a standard object detector~\cite{icpr2018,su2017watchable,yu2018deep}. However, this approach makes it difficult to detect objects located on the boundary of each perspective sample while performing individual object detection on overlapping multiple perspective images leads to a computational cost problem. The other method directly applies a perspective object detector on equirectangular projection (ERP) images~\cite{icvrv2017} where 360\textdegree~information is projected to the plane, however mapping a spherical image of a view to a planar image inevitably introduces projective distortion, thereby increasing the difficulty of object detection. 


Some object detection models have been designed for 360\textdegree~images based on the concept of spherical convolution (SphConv) ~\cite{su2017,coors2018spherenet} where the CNN kernel is adapted to a spherical surface.
Su et al.~\cite{su2017} proposed SphConv which uses different kernels for each row and increasing the kernel size towards the polar regions. Coors et al.~\cite{coors2018spherenet} proposed SphereNet to adjust the sampling grid location of kernels according to the equirectangular projection, and learn spherical image representations by encoding distortion into convolutional filters. However, those methods are hard to be applied to high-resolution images or deep models due to their inefficiency. Furthermore, the complex spherical network architecture suffers from integrating the techniques in state-of-the-art perspective object detectors. Therefore, we need a framework that can be directly applied to the convenient ERP format but robust to the projective distortion.

For handling the distortion in the ERP format, a better design of the bounding box (BB) and the Intersection-over-Union (IoU) in the 360\textdegree~image is important. BB and IoU are two fundamental elements of object detection where bounding box serves the spatial location and boundary of an object while IoU represents the degree of overlap of two bounding boxes (i.e., ground-truth and prediction). While IoU is generally used as an evaluation metric, Non-maximum Suppression (NMS) at inference, which is a technique to suppress redundant predictions of object bounding boxes, is also based on the IoU score. In addition, some researches also use IoU as a part of loss function~\cite{iouloss,giou} for bounding box regression. 

In the 360\textdegree~object detection scenario, the usual rectangular BB in 2D image coordinates cannot properly constrain objects on a sphere especially near the poles, therefore it is common to use Field-of-View Bounding Box (FoV-BB, a.k.a spherical bounding boxes)~\cite{icpr2018, zhao2020criteria} which is defined with the center location and left-right/up-down FoVs of the object’s occupation. While FoV-BB is specifically defined for 360\textdegree~images, it is nontrivial to compute the area of intersection between two FoV-BBs with different centers due to their non-rectangular shape. To tackle this challenge, SphericalCriteria~\cite{zhao2020criteria} has recently proposed SphericalIoU (Sph-IoU) to approximate the IoU between two FoV-BB as a decision criterion for 360\textdegree~object detection. Concretely, Sph-IoU undistorted FoV-BBs by moving them to the equator along the longitude, then calculated their intersection as they are the rectangular BBs (See Fig.~\ref{fig:sph_fov}). While this approach works well when FoV-BB is in relatively low latitude regions, Sph-IoU largely underestimates the area of the original intersected area since it regards the longitude difference as the horizontal distance between centers of two bounding boxes. In other words, moving centers of FoV-BBs to the equator changes the great-circle distance between two centers which should be preserved.

In this paper, we propose new technical components that are applicable to any object detectors for 360\textdegree~images in the ERP format. To address the underestimation problem of Sph-IoU, we propose a novel IoU computation method, namely {\it FoV-IoU} which better approximates the exact computation of IoU between two FoV-BBs (\textbf{contribution 1}). In addition, we also propose {\it 360Augmentation}, a 360\textdegree~image augmentation technique that can increase the diversity of training data while maintaining the spherical coordinate mapping of the ERP image (\textbf{contribution 2}). The proposed method is compared with various 360-degree object detection methods on the 360-Indoor dataset~\cite{chou2020indoor}, and is shown to perform well in combination with various existing perspective object detectors (\textbf{contribution 3}).

The remainder of this paper is organized as follows. In Section II, we introduce prior knowledge and problems. In Section III, we present the Field-of-View IoU and its advantages. In Section IV, we present 360Augmentation. Experimental results and corresponding analysis are provided in Section V. In Section VI, we conclude this paper.

\section{Background}
\subsection{Field-of-View Bounding Box (FoV-BB)} 
Bounding Box (BB) is usually used to locate objects and evaluate the performance in the object detection task. For perspective images, rectangular bounding boxes are commonly represented in the form of $(x,y,w,h)$, where $x$ and $y$ are the coordinates of the center point and $w$ and $h$ are the width and height of a bounding box, respectively. However, the conventional bounding box cannot tightly bound an object in an ERP image since the objects are heavily distorted. In other words, the rectangular BB in an ERP image is distorted when being projected onto the perspective view, especially in upper and lower latitude regions. Therefore, some researches and datasets of 360\textdegree~object detection~\cite{zhao2020criteria,chou2020indoor} use Field-of-View Bounding Box (FoV-BB) instead of the normal bounding box. 

Field-of-View (FoV) typically refers to the widest observable scene of a camera, and it is also used to define the size of an object in a 360\textdegree~image. The FoV-BB uses vertical and horizontal FoV angles instead of pixel coordinates. A Field-of-View bounding box can be defined as $(\theta, \phi, \alpha, \beta)$ where $\theta$ and $\phi$ are the longitude and latitude coordinates of the box center, and $\alpha$ and $\beta$ are the horizontal and vertical FoVs which represent the occupation of an object.

\subsection{Difficulty of IoU computation for FoV-BB}
Different from normal bounding boxes, the overlap of FoV-BBs is not rectangular, which makes it difficult to calculate the area directly. The most unbiased way to calculate the IoU of two FoV-BBs is to use spherical polygons~\cite{sphpoly}, which are defined by a number of intersecting great-cycles on the sphere as follow:
\begin{equation}
A(Spherical Polygon) = \{\Sigma-(n-2) \pi\} r^{2}.
\end{equation}
Here $A$ means Area, $n$ is the number of sides (edges) of the polygon, $\Sigma$ is the sum of all its angles and $r$ is the radius of the sphere. 

However, calculating IoU based on spherical polygons is not practical for the object detection task. Unlike the shape of the intersection of normal BBs, one of FoV-BBs is irregular, and the calculation of each interior angle of a spherical polygon requires multiple trigonometric function calculations, which is too inefficient. 
In addition to performance issues, spherical polygon-based IoU is not differentiable, therefore cannot benefit from introducing IoU-based loss which has recently been claimed to be useful for object detection. Therefore, we want to find an efficient and differentiable approximation method for computing the intersection of FoV-BBs.

\subsection{Problem of Sph-IoU} \label{sec:Sph-IoU}
Zhao et al.~\cite{zhao2020criteria} have recently proposed Sph-IoU, which is an approximated IoU between two FoV-BBs. 
Sph-IoU regards FoV-BB (called Sph-BB in~\cite{zhao2020criteria}) as the proportion of a spherical segment. For a FoV-BB $(\theta, \phi, \alpha, \beta)$, Sph-IoU calculates its area by:
\begin{gather}
A({\rm FoV\mathchar`-BB}) = 2\alpha\sin(\beta/2).
\end{gather}
Although the proportion of a spherical segment is not equivalent to spherical polygons, this approximation is acceptable.
For computing the area of intersection of two FoV-BBs, Sph-IoU simply replaced planar coordinates by FoV coordinates as follows:
\begin{gather}
\theta_{{\rm min}}^I = \mathrm{{\rm max}}( \theta_g - \frac{\alpha_g}{2}, \theta_d - \frac{\alpha_d}{2}),\nonumber \\
\theta_{{\rm max}}^I = \mathrm{{\rm min}}(\theta_g + \frac{\alpha_g}{2}, \theta_d + \frac{\alpha_d}{2}),\nonumber \\
\phi_{{\rm min}}^I = \mathrm{{\rm max}}( \phi_g - \frac{\beta_g}{2}, \phi_d - \frac{\beta_d}{2}),\nonumber \\
\phi_{{\rm max}}^I = \mathrm{{\rm min}}(\phi_g + \frac{\beta_g}{2}, \phi_d + \frac{\beta_d}{2}), \nonumber \\ 
A(I) = (\theta_{{\rm max}}^I - \theta_{{\rm min}}^I) \times (\phi_{{\rm max}}^I - \phi_{{\rm min}}^I),
\label{eq:Sph-IoU}
\end{gather}
where $g$ is the subscript for the ground truth bounding box and $d$ is one for the detected. As~\cite{zhao2020criteria} mentioned, this computation is equivalent to moving two BBs to the equator along the longitude, then computing the rectangular-like intersection area as illustrated in Fig.~\ref{fig:sph_fov}. 

However, Sph-IoU has a serious problem with this approximation. Moving BBs onto the equator along the longitude line inevitably increases the great-cycle distance between two FoV-BBs on a sphere, thus underestimates the area of intersection, especially for those objects located at high latitudes originally. 
\section{Method}
In this section, we introduce Field-of-View IoU (FoV-IoU), an efficient, accurate, and differentiable IoU computation method between two FoV-BBs in an ERP image. We first explain the Field-of-View IoU computation process, then demonstrate the benefit of using FoV-IoU in training and testing.

\subsection{Field-of-View IoU (FoV-IoU)}
The main idea of FoV-IoU is the rectangular approximation of the irregular intersection area of two FoV-BBs similarly with Sph-IoU but using the great-cycle distance on the sphere between centers of FoV-BBs instead of using the azimuth difference. 

Generally, IoU computation requires the areas of bounding boxes $A(B_d)$, $A(B_g)$ and their intersection $A(I)$. Remind that the intersection area of normal BBs is a rectangle based on the positions of the four sides of two BBs. This area could be computed from the distance between the centers of two BBs and the width and height of each BB. However, the accurate and efficient computation of $A(I)$ for FoV-BBs is non-trivial as has already been discussed. As shown in Section~\ref{sec:Sph-IoU}, Sph-IoU takes the difference in longitude and latitude as the distance between the centers of two bounding boxes, but this approximation is suboptimal which always overestimates the great-cycle distance of centers at high-latitude regions and thus makes IoU underestimated. 

Unlike Sph-IoU which uses longitude/latitude differences to compute the IoU and underestimate the distance between two BBs, FoV-IoU computes the distance more accurately by using the great-circle distance, which is the shortest distance between two points on a sphere.  In general, given the spherical coordinates of two points, the great-circle distance can be calculated by Haversine formula~\cite{hav}. Although Haversine formula calculates great-circle distance accurately, it is still inefficient in practical. We can simplify this calculation by discomposing great-circle distance at two directions, and using the equirectangular projection formula: 
\begin{align}
    x &= R \times (\theta-\theta_0) \times  \cos (\phi_{sp}),  \\
    y &= R \times (\phi-\phi_0),
\end{align}
where ($x$, $y$) is the projected coordinate on ERP, ($\theta$, $\phi$) is longitude and latitude coordinates on sphere, ($\theta_0,\phi_0$) is the center coordinate of ERP and $\phi_{sp}$ is the standard parallel which is the line with no distortion on ERP. Generally, $\phi_{sp}$ is zero in an ERP image, which means there is no distortion on the equator. For vertical direction, the great-circle distance is linear to the latitude difference in the ERP coordinate. For horizontal difference, we calculate the approximate horizontal great-circle distance, namely {\it FoV distance} ($\Delta fov$) between center points of $B_g$ and $B_d$ as
\begin{align}
\Delta fov = (\theta_d-\theta_g)\times \cos \left(\frac{\phi_g+\phi_d}{2}\right).
\end{align}
Intuitively speaking, we consider different ERP projections for every pair of FoV-BBs where their standard parallels are represented as $\phi_{sp}=\frac{\phi_g+\phi_d}{2}$. $\Delta fov$ is equivalent to projecting the middle of two center points to the equator of the new ERP coordinate then calculate the horizontal distance between two center points in this ERP coordinate. Since there is no distortion at the latitude of the midpoint of the two centers of FoV-BBs, the longitude distance approximates the horizontal great-circular distance.
\begin{algorithm}[t]
\caption{FoV-IoU Calculation}
\hspace*{0.02in} {\bf Input:}
$B_g = (\theta_g, \phi_g, \alpha_g, \beta_g)$, $B_d = (\theta_d, \phi_d, \alpha_d, \beta_d)$, where $B_g$ is a ground truth bounding box and $B_d$ is a detected bounding box.\\
\hspace*{0.02in} {\bf Output:} 
IoU between $B_g$ and $B_d$ \\

1. Compute FoV Area of $B_g$ and $B_d$: \\
$A(B_g)=\alpha_g\times \beta_g$, $A(B_d)=\alpha_d\times \beta_d$. \\

2. Compute FoV distance between $B_g$ and $B_d$: \\
$\Delta fov = (\theta_d-\theta_g)\times \cos(\frac{\phi_g+\phi_d}{2})$. \\

3. Build an approximate FoV intersection $I$: \\
$\theta_{{\rm min}}^I = \mathrm{{\rm max}}( - \frac{\alpha_g}{2},\, \Delta fov - \frac{\alpha_d}{2})$, \\
$\theta_{{\rm max}}^I = \mathrm{{\rm min}}(\frac{\alpha_g}{2},\, \Delta fov + \frac{\alpha_d}{2})$, \\
$\phi_{{\rm min}}^I = \mathrm{{\rm max}}(\phi_g - \frac{\beta_g}{2},\, \phi_d - \frac{\beta_d}{2})$, \\
$\phi_{{\rm max}}^I = \mathrm{{\rm min}}(\phi_g + \frac{\beta_g}{2},\, \phi_d + \frac{\beta_d}{2})$. \\

4. Compute area of FoV intersection $I$ and union $U$:  \\
$A(I) = (\theta_{{\rm max}}^I - \theta_{{\rm min}}^I) \times (\phi_{{\rm max}}^I - \phi_{{\rm min}}^I)$, \\
$A(U) = A(B_g)+A(B_d)-A(I)$.\\

5. Compute FoV-IoU by general IoU form: \\
$\mathrm{FoVIoU}(B_g, B_d) = \frac{A(I)}{A(U)}$. \\
\label{algo}
\end{algorithm}

The detailed FoV-IoU computation process is shown in Alg.~\ref{algo}. Assume we have ground truth and detected bounding boxes $B_g$ and $B_d$, we first calculate the areas of $A(B_d)$, $A(B_g)$ by multiplying $\alpha_{d,g}$ and $\beta_{d,g}$. Then, we use $\Delta fov$ and $\alpha_{d,g}, \beta_{d,g}$ to compute $\theta_{{\rm max}}^I$ and $\theta_{{\rm min}}^I$ which are supposed to be the left-right boundaries of intersecting areas. Concretely, we simply replace $\theta_g$ by zero and $\theta_d$ by $\Delta fov$ in E.q.(\ref{eq:Sph-IoU}). We also compute $\phi_{{\rm max}}^I$ and $\phi_{{\rm min}}^I$ from the latitude coordinates of FoV-BBs' centers and their FoVs. Then, we compute the approximated FoV intersection $A(I)$ by ($\theta_{{\rm max}}$,  $\theta_{{\rm min}}$, $\phi_{{\rm max}}$, $\phi_{{\rm min}}$), from the analogy of normal IoU definition ($x_{{\rm max}}$, $x_{{\rm min}}$, $y_{{\rm max}}$, $y_{{\rm min}}$). Given $A(B_d)$, $A(B_g)$ and $A(I)$, we compute the FoV-IoU score just by following the IoU computation process as in E.q.(\ref{eq:Sph-IoU}).

\begin{table}[]
    \centering
    \caption{Comparison of different types of IoU for Field-of-view bounding box. The blue(left) box's coordinate is (30°, 60°, 60°, 60°) and red(right) box's coordinate is (60°, 60°, 60°, 60°). GT IoU is the exact IoU calculated based on spherical polygon.}
    \begin{tabular}{m{1cm}<{\centering}|m{2cm}<{\centering}|m{2cm}<{\centering}|m{2cm}<{\centering}}
    \toprule
    Type & Exact IoU & FoV-IoU & Sph-IoU \\
    \hline       
    Approx. pattern & \includegraphics[width=0.25\columnwidth]{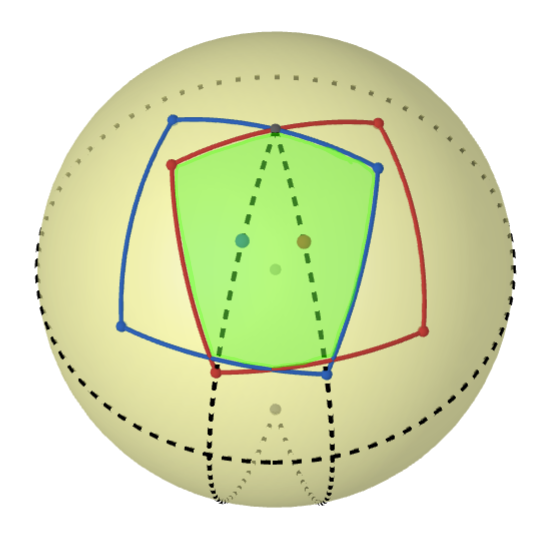} &  \includegraphics[width=0.25\columnwidth]{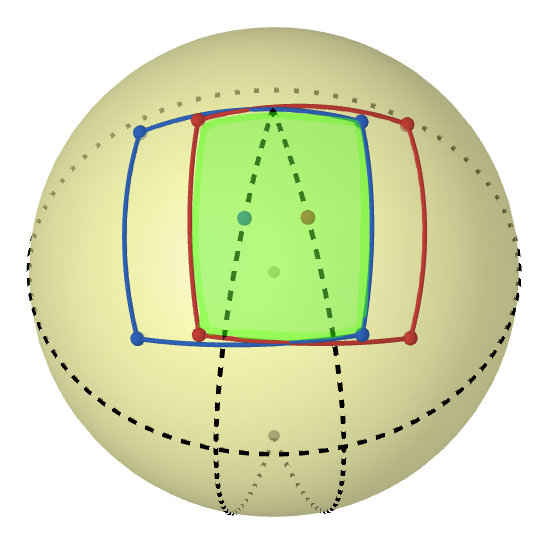} & \includegraphics[width=0.25\columnwidth]{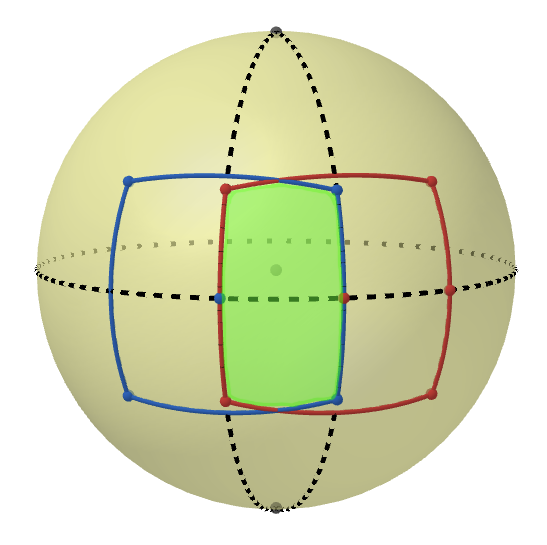}  \\
    \hline
    IoU & 0.57 & 0.59 & 0.33 \\
    \bottomrule
    \end{tabular}
    \label{tab:compare_three_ious}
\end{table}

\subsection{Advantages of FoV-IoU}  \label{sec:advantages}
IoU plays an important role in evaluation, inference and training. Here we illustrate the benefits of FoV-IoU for each stage respectively.

\subsubsection{Evaluation}
Mean Average Precision (mAP) is a popular evaluation metric for object detection, which represents the ratio of False Positive (FP) and True Positive (TP). In mAP calculation, IoU score is used to determine whether a prediction is FP or TP based on the IoU threshold. For evaluation, Sph-IoU significantly underestimates the overlapped area, especially for high-latitude objects. On the other hand, FoV-IoU is consistent in different latitudes and the score is more identical to the exact IoU. We illustrate an example in Table~\ref{tab:compare_three_ious}. Assuming that the blue box is the ground truth and the red box is the predicted box respectively, we can see that the approximate intersection by FoV-IoU is much closer to one by the exact IoU than one by Sph-IoU. Quantitatively speaking, Sph-IoU largely underestimated the IoU value from 0.57 to 0.33, thus the prediction wasn't be selected as TP with 0.5 IoU threshold while our FoV-IoU accurately approximated the exact IoU in this example.

\subsubsection{Inference} 
Non-maximum Suppression (NMS) is a post-processing component for filtering out redundant predictions at inference stage based on the IoU score. Underestimated Sph-IoU score can lead to more overlapping predictions and due to the complexity of FoV-BB, it is impractical to calculate the area of the spherical polygon (exact IoU) in the detection network. On the other hand, using FoV-IoU allows efficient and accurate calculation of NMS in the inference stage, resulting in better prediction results.

\subsubsection{Training} We integrate FoV-IoU with Generalized IoU (GIoU) loss for bounding box regression. IoU-based losses~\cite{iouloss}~\cite{giou} are proposed to eliminate the gap between training and testing, which has been widely used in object detection. Inspired by the idea of IoU-based loss, we implemented {\it FoV-GIoU loss} where IoU computation in the GIoU loss is replaced by our FoV-IoU. The FoV-GIoU loss is explained in Alg.~\ref{alg:loss}. Similar to the intersection, we calculate the area of the smallest enclosing box according to the horizontal FoV distance $\Delta fov$.

\begin{algorithm}[h]
\caption{FoV-GIoU Loss}
\hspace*{0.02in} {\bf Input:}
$B_g = (\theta_g, \phi_g, \alpha_g, \beta_g)$, $B_d = (\theta_d, \phi_d, \alpha_d, \beta_d)$, where $B_g$ is a ground truth bounding box and $B_d$ is a detected bounding box.\\
\hspace*{0.02in} {\bf Output:} 
$\mathcal{L}_{FoVIoU}$. \\

1. Compute $\Delta fov$, FoV intersection $I$ and FoV union $U$ as Alg.~\ref{algo}. \\

2. Build an approximate smallest enclosing box $C$ ~\cite{giou}: \\
$\theta_{{\rm min}}^C = \mathrm{{\rm min}}( - \frac{\alpha_g}{2},\, \Delta fov - \frac{\alpha_d}{2})$,\\
$\theta_{{\rm max}}^C = \mathrm{{\rm max}}(\frac{\alpha_g}{2},\, \Delta fov + \frac{\alpha_d}{2}),$ \\
$\phi_{{\rm min}}^C = \mathrm{{\rm min}}(\phi_g - \frac{\beta_g}{2},\, \phi_d - \frac{\beta_d}{2})$,\\
$\phi_{{\rm max}}^C = \mathrm{{\rm max}}(\phi_g + \frac{\beta_g}{2},\, \phi_d + \frac{\beta_d}{2}).$ \\

3. Compute area of smallest enclosing box $C$:  \\
$A(C) = (\theta_{{\rm max}}^C - \theta_{{\rm min}}^C) \times (\phi_{{\rm max}}^C - \phi_{{\rm min}}^C)$. \\

4. $\mathcal{L}_{FoVIoU} = 1 - \mathrm{FoVIoU}(B_g, B_d) + \frac{A(C) - A(U)}{A(C)}$. \\
\label{alg:loss}
\end{algorithm}
FoV-GIoU loss is proposed to solve the imbalance problem at different latitudes. As a comparison, we implemented Sph-GIoU loss which integrates Sph-IoU to GIoU loss in the following experiment. Both normal L1 loss and Sph-GIoU loss are uniform on the ERP image but not uniform on the sphere surface. As a result, the loss at high latitudes is more difficult to be optimized. We consider that FoV-GIoU loss can address this problem by paying attention to position relation on the sphere.

\begin{figure*}[t]
    \centering
    \includegraphics[width=1.0\linewidth]{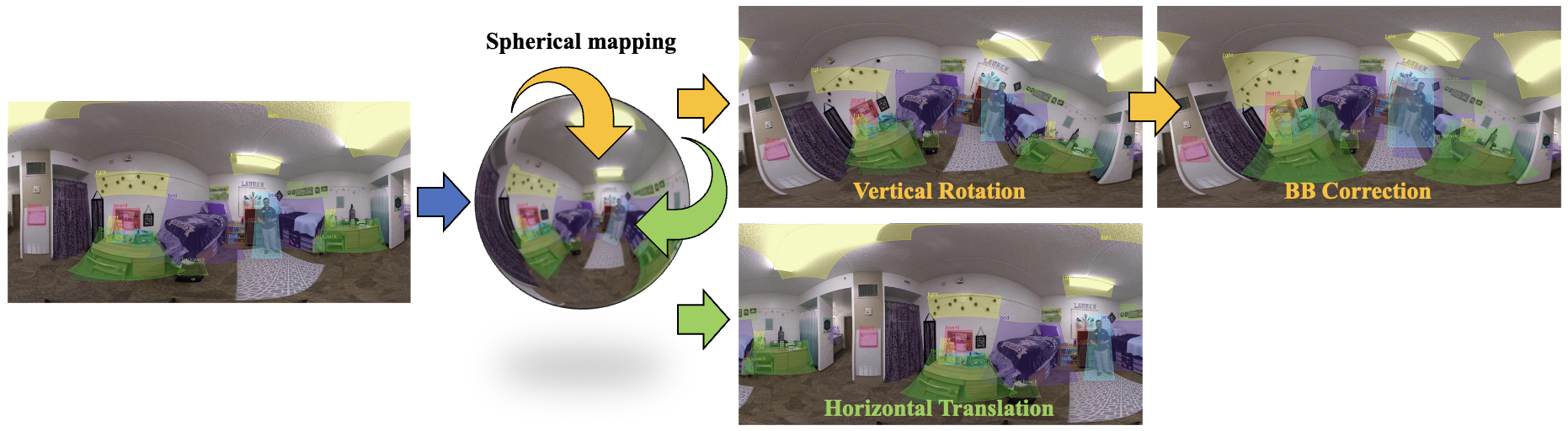}
    \caption{The visualized pipeline of 360Augmentation. We randomly applied Horizontal Translation and Vertical Rotation on the training sets. Additionally, we resize the BB after Vertical Rotation. The experimental results show that 360Augmentation can consistently improve the accuracy of object detection in 360\textdegree~images.}
    \label{fig:aug_full}
\end{figure*}

\section{360Augmentation}
Many data augmentation strategies have been proposed for perspective images. Geometric transformations such as rotation and translation are an important part of image data augmentation. However, augmentation methods for perspective geometric transformation are often unsuitable for ERP images. Because there is a fixed mapping between an ERP image and a sphere, simply cropping or rotating the ERP image will destroy the spherical surface mapping, resulting in inconsistency between training and inference data. In addition, in the context of object detection, bounding boxes must be augmented according to the augmentation of the image. Thus, we propose {\it 360Augmentation}, a data augmentation technique for the 360\textdegree~object detection which can increase the diversity of training data while maintaining the spherical coordinate mapping of the ERP image. To the best of our knowledge, this is the first argumentation method specifically for 360\textdegree~object detection task.

360Augmentation includes two strategies: vertical random rotation and horizontal translation. The augmentation pipeline is shown in Fig.~\ref{fig:aug_full}. The main idea of our augmentation method is to imitate human visual perception in 3D. When we use a virtual reality (VR) device to watch a 360\textdegree~scene, we often turn our heads up and down or turn around, to observe the objects around us. However, we do not turn the view upside down or tilt the view. Based on this observation, we rotate the image without violating the natural human perception. Concretely, we first convert latitude and longitude coordinates $(\phi, \theta)$ to Cartesian coordinates $(x, y, z)$ for every pixel by spherical projection as
\begin{align}
(x,\, y,\, z) = (\sin{\theta} \cos{\phi},\, \sin{\phi},\,  \cos{\theta} \cos{\phi}).
\end{align}
We then apply horizontal translation and vertical rotation randomly. For horizontal translation, we randomly select translation angle from 0\textdegree~to 360\textdegree~, then accordingly shift the image and bounding box by applying transformation on $(x, y, z)$ coordinates as
\begin{align}
(x^\prime,\, y^\prime,\, z^\prime) = (x\cos{\theta_r}-z\sin{\theta_r},\, y,\, x\sin{\theta_r}+z\cos{\theta_r}),
\end{align}
where $\theta_r$ is randomly selected from 0\textdegree~ to 360\textdegree~. We applied vertical random rotation to improve the detection of high-latitude objects. We rotate both the image and the center point of bounding boxes by:
\begin{align}
(x^\prime,\, y^\prime,\, z^\prime) = (x,\, y \cos{\phi_r}+z\sin{\phi_r},\, -y\sin{\phi_r}+z\cos{\phi_r}),
\end{align}
where $\phi_r$ is a random angle chosen from $-30^{\circ}$ to $30^{\circ}$. 

A challenge of rotation augmentation is determining the size of an augmented bounding box. In the case of a perspective image, we can first rotate boxes as same as the image, then define the tightest parallel bounding box fit to the rotated bounding box as an augmented annotation. Assume $(w, h)$ are width and height for original annotation and $\delta$ is the rotation angle, then width and height of augmented annotation can be calculated by:
\begin{align}
(w',h') = (w \cos{\delta} + h \sin{\delta}, w \sin{\delta} + h \cos{\delta}).
\end{align}
In case of an ERP image, rotation angle is different at each position in the ERP image. To rotate bounding box in it, we first map the center point of the bounding box to its new coordinates, then compute the rotation angle at the center point, and finally rotate and resize the bounding box.

We consider that vertical random rotation can reduce the distortion of the objects in the upper and lower regions, thereby making recognition easier. After translation and rotation, we remap Cartesian coordinates back to the latitude and longitude to generate the augmented images.

\section{Experiments}

We evaluated the performance of our method by combining existing perspective object detection models with Sph-IoU/FoV-IoU and 360Augmentation, and comparing them against other 360\textdegree~object detection models. 
\subsection{Implementation details}
\subsubsection{Datasets} Our experiments were conducted on the recent 360-Indoor dataset~\cite{chou2020indoor}. Real benchmark datasets for the 360\textdegree~object detection task are quite limited compared to ones for general perspective images. In reality, before 360-Indoor was presented, evaluations were made with synthetic data alone~\cite{coors2018spherenet,zhao2020criteria}, which did not reflect the complex scenes of the real world. 360-Indoor is the largest 360\textdegree~object detection dataset, which contains 3335 real-world indoor 360\textdegree~images with high-resolution ($1920\times960$) and 89148 bounding boxes of 37 categories that locate from low latitude regions to high latitude regions. We used 2k images for training and 1k images for testing. Note that the annotations of BBs in the 360-Indoor dataset are defined by FoV-BB. 

\subsubsection{Evaluation protocol}  Methods were evaluated with the same evaluation metrics as MS COCO dataset~\cite{coco} including AP with different IoU threshold (0.50, 0.55, ..., 0.95). Furthermore, we calculated high-latitude AP for better evaluating the effect of the proposed method for the high-latitude region in 360\textdegree~images. AP values are calculated based on FoV-IoU to adapt to the FoV-BB and produce an accurate result. 
\subsubsection{Training details}  All experiments were conducted with mmdetection implementations~\cite{chen2019mmdetection} except for YOLOv3 PyTorch implementation released by Ultralytics LLC~\cite{ultralytics}. Microsoft COCO dataset~\cite{coco} pre-trained model is used for the parameter initialization. All of the models were trained with a single GPU (Quadro RTX 8000 or A100-PCIE-40GB) for 50 epochs. For a fair comparison, We used the default parameter settings in mmdetection and disabled geometric data augmentation methods, and the same image resolution (i.e., [1920, 960]) for training and test in all following experiments.

\subsection{FoV-IoU vs Sph-IoU}
\subsubsection{IoU Accuracy and Efficiency}

\begin{table*}[t]
    \centering
    \caption{More example of IoU of FoV-BBs. We visualize IoU between b1(blue box) and b2(red box) by both ERP and PSP (box-shadowed) views. The green areas are the overlap areas between b1 and b2.}
    \begin{tabular}{c|c|c|c}
    \toprule
    Coords & \makecell{
    b1 = [40\textdegree, 50\textdegree, 35\textdegree, 55\textdegree] \\ 
    b2 = [35\textdegree, 20\textdegree, 37\textdegree, 50\textdegree]}& 
    \makecell{
    b1 = [30\textdegree, 60\textdegree, 60\textdegree, 60\textdegree] \\ 
    b2 = [55\textdegree, 40\textdegree, 60\textdegree, 60\textdegree]}& 
    \makecell{
    b1 = [50\textdegree, -78\textdegree, 25\textdegree, 46\textdegree] \\ 
    b2 = [30\textdegree, -75\textdegree, 26\textdegree, 45\textdegree]} \\
    \hline       
    ERP \& PSP view & 
    \raisebox{-.5\height}{\includegraphics[width=0.25\linewidth]{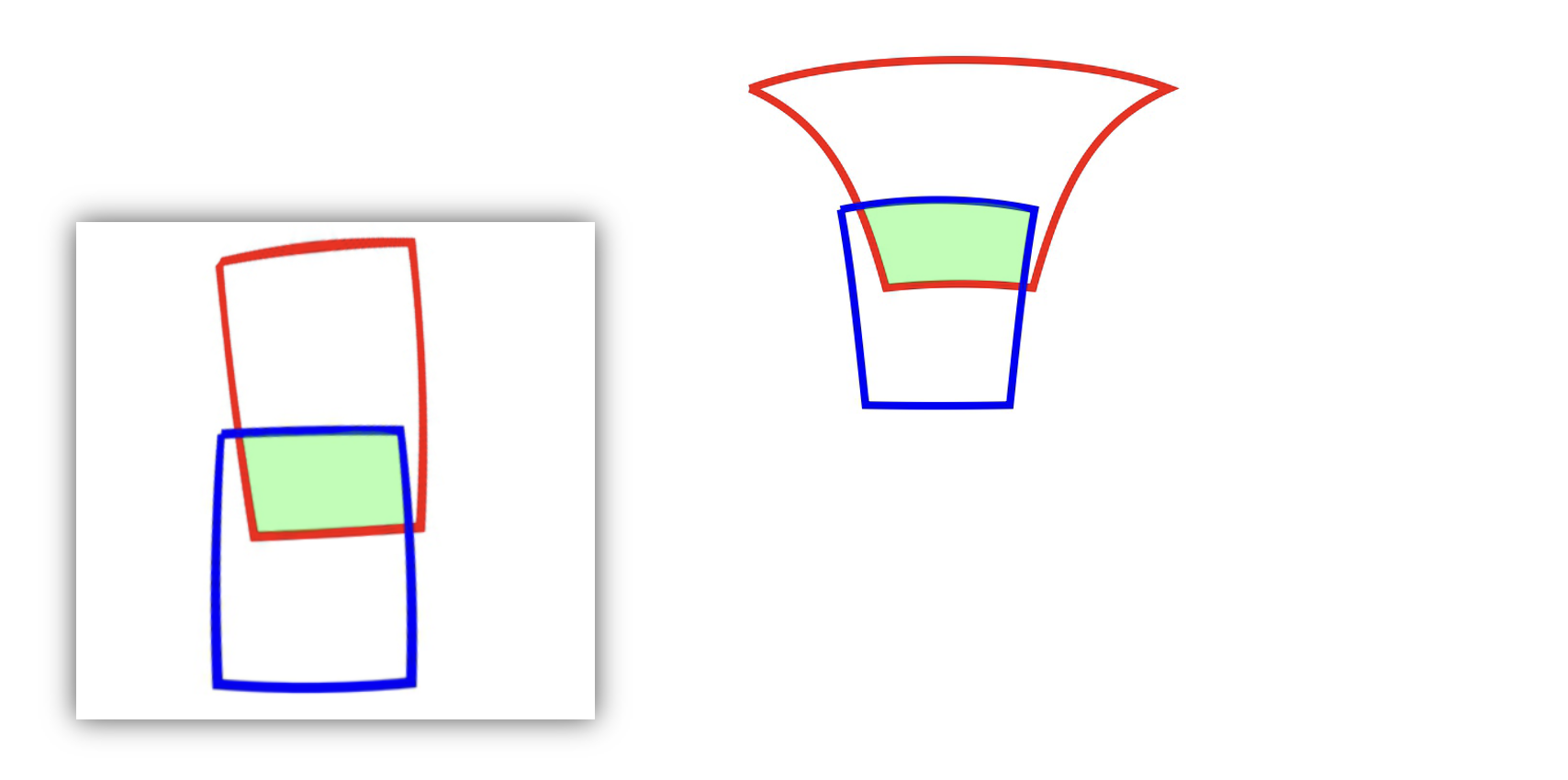}} & \raisebox{-.5\height}{\includegraphics[width=0.25\linewidth]{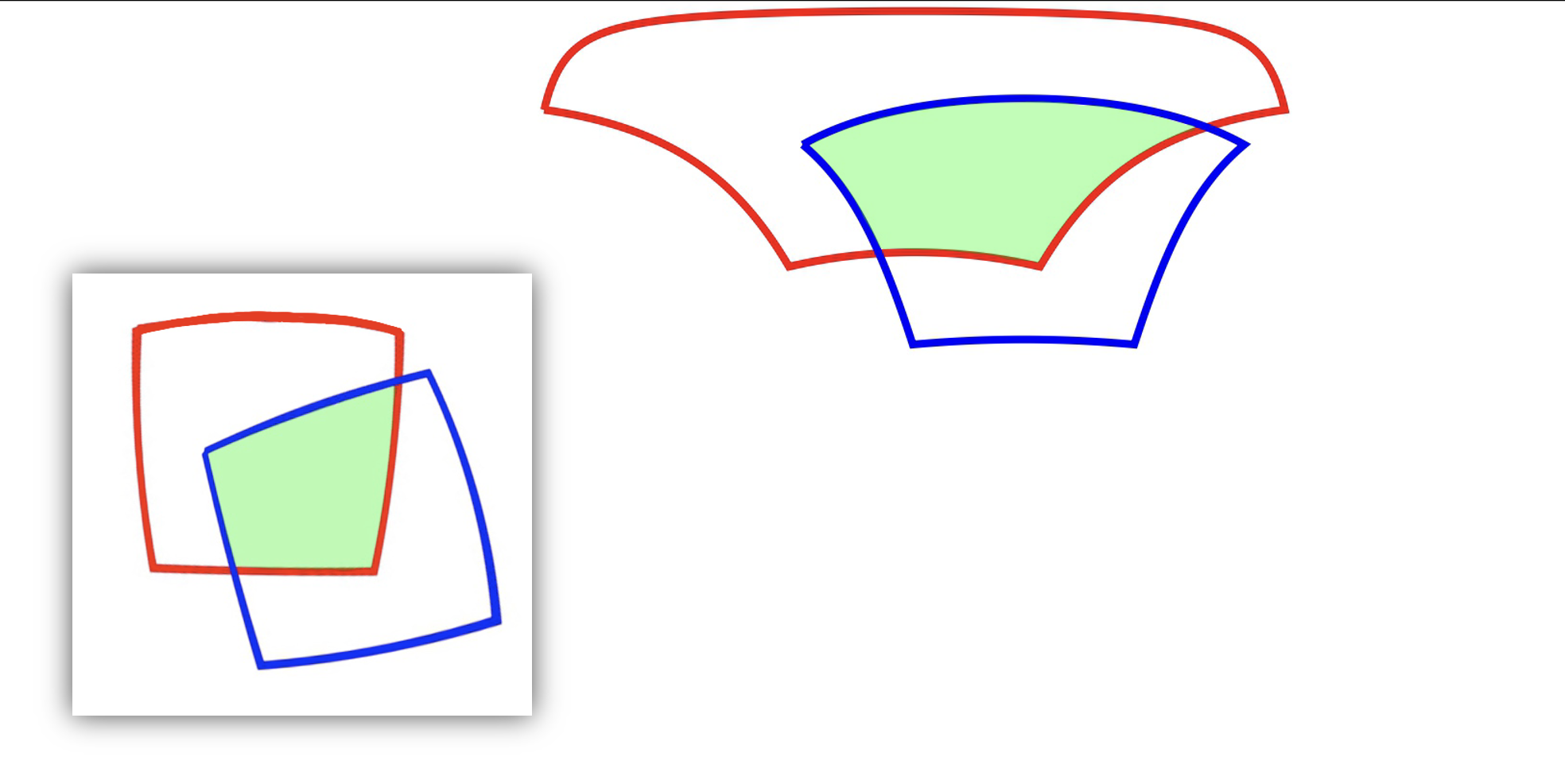}} & \raisebox{-.5\height}{\includegraphics[width=0.25\linewidth]{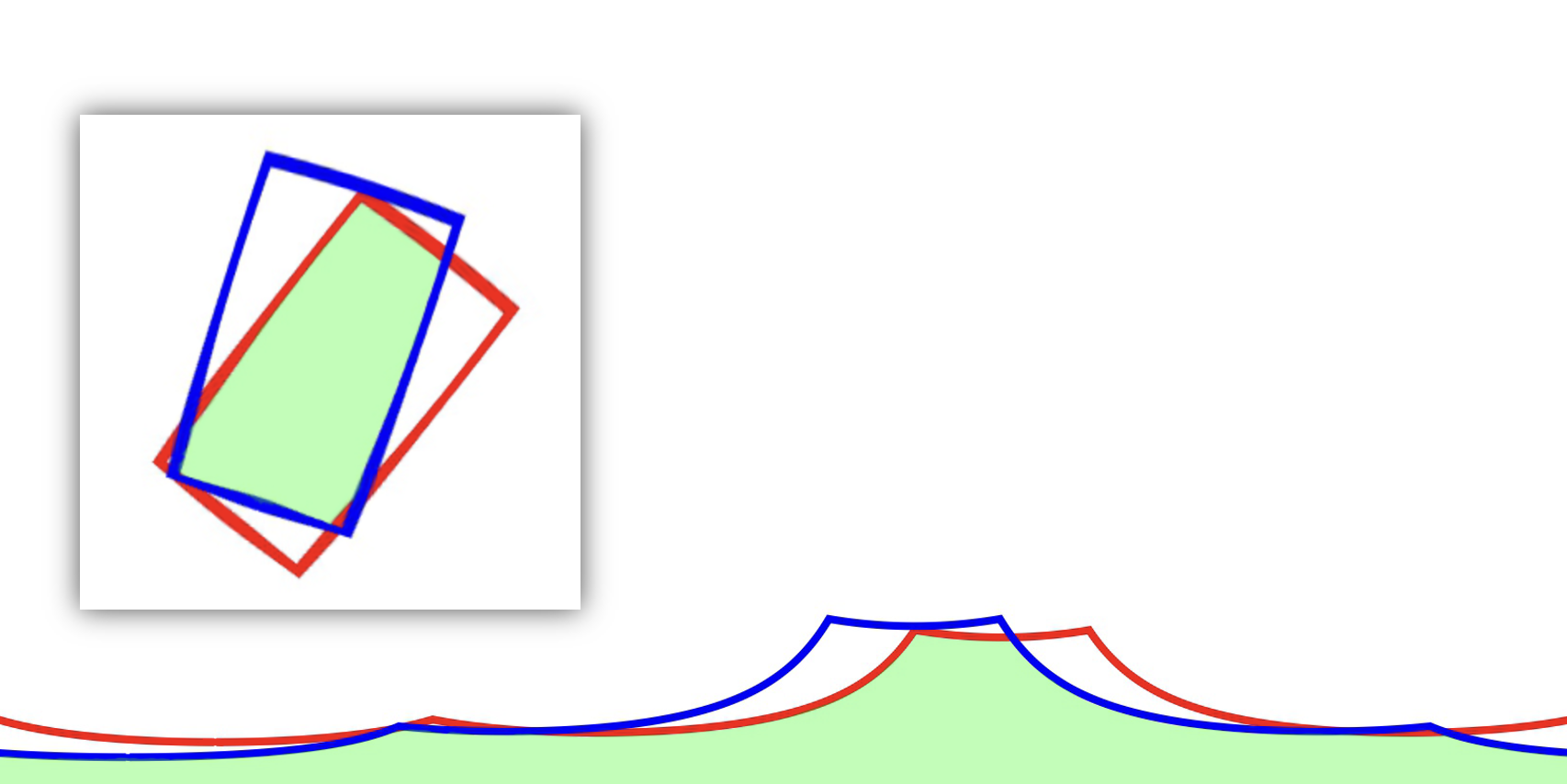}} \\
    \hline  
     Sph-IoU & 0.227 (-0.021) & 0.250 (-0.075) & 0.112 (-0.515) \\
     \hline  
     FoV-IoU & 0.235 (-0.013) & 0.323 (-0.002) & 0.617 (-0.010) \\
     \hline  
     Exact IoU & 0.248 & 0.325 & 0.627 \\

    \toprule
    Coords & \makecell{
    b1 = [30\textdegree, 75\textdegree, 30\textdegree, 60\textdegree] \\
    b2 = [60\textdegree, 40\textdegree, 60\textdegree, 60\textdegree]}& 
    \makecell{
    b1 = [40\textdegree, 70\textdegree, 25\textdegree, 30\textdegree] \\ 
    b2 = [60\textdegree, 85\textdegree, 30\textdegree, 30\textdegree]} & 
    \makecell{
    b1 = [30\textdegree, 75\textdegree, 30\textdegree, 30\textdegree] \\ 
    b2 = [60\textdegree, 55\textdegree, 40\textdegree, 50\textdegree]} \\
    \hline       
    ERP \& PSP view & 
    \raisebox{-.5\height}{\includegraphics[width=0.25\linewidth]{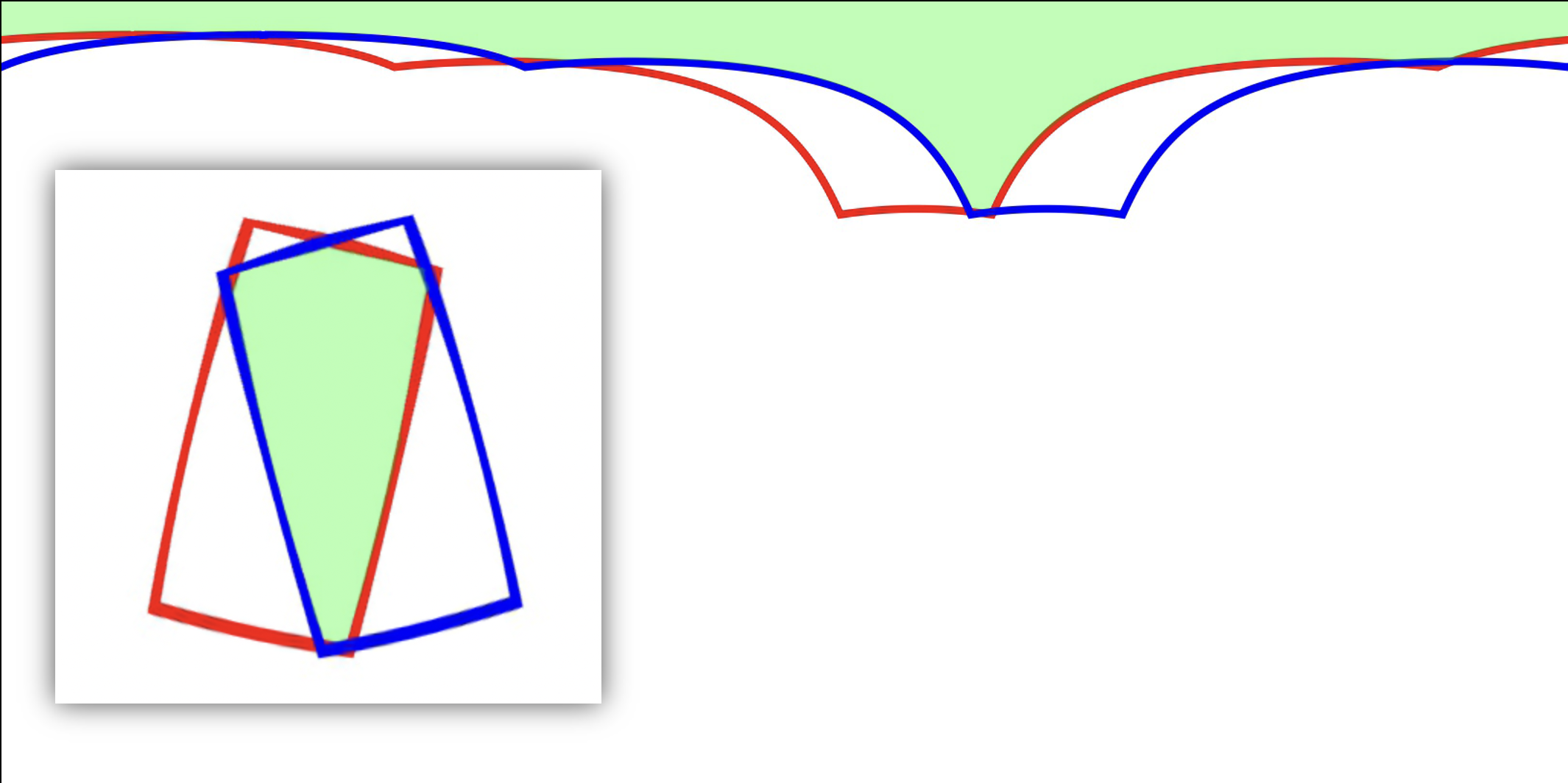}} & \raisebox{-.5\height}{\includegraphics[width=0.25\linewidth]{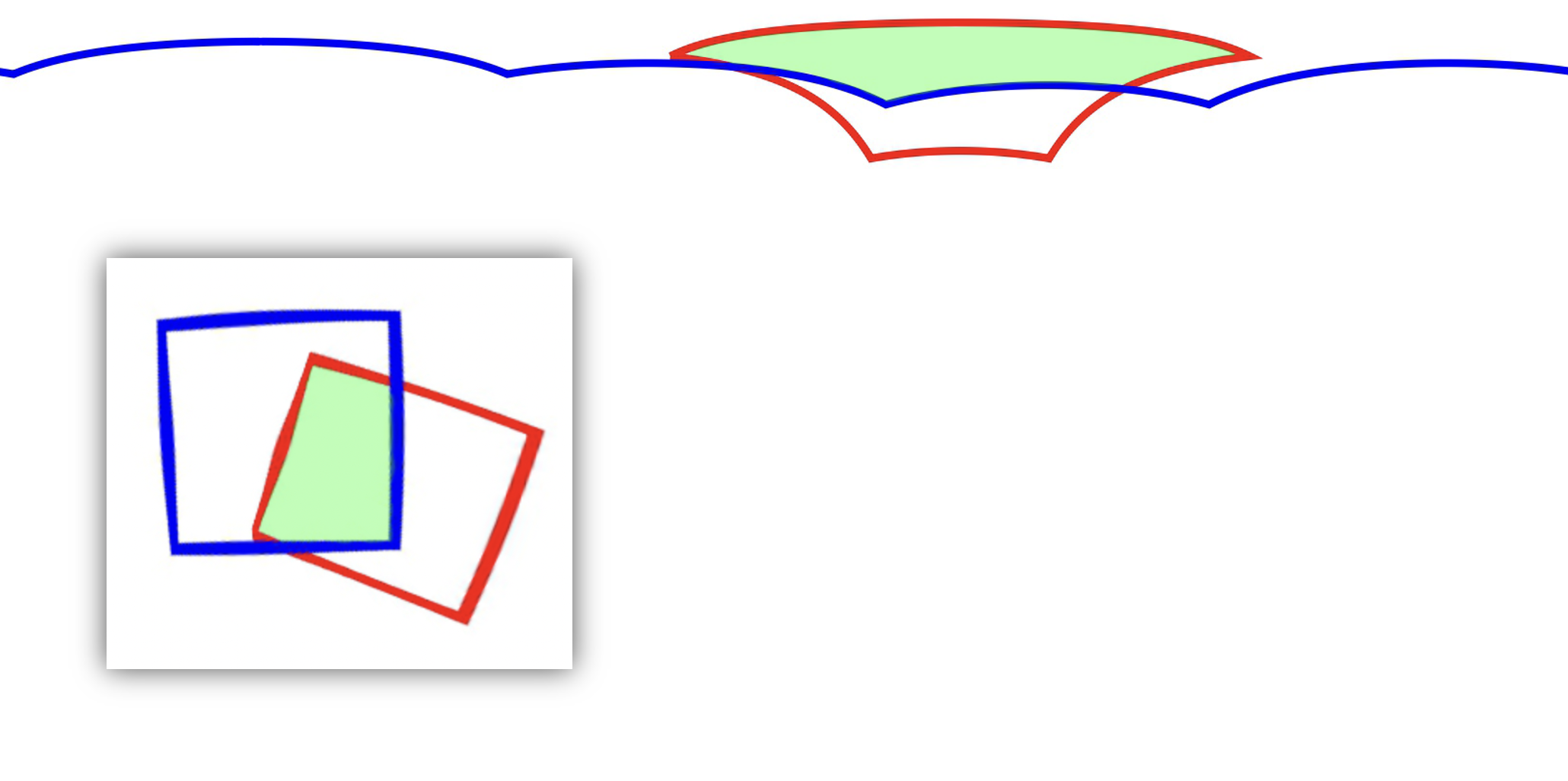}} & \raisebox{-.5\height}{\includegraphics[width=0.25\linewidth]{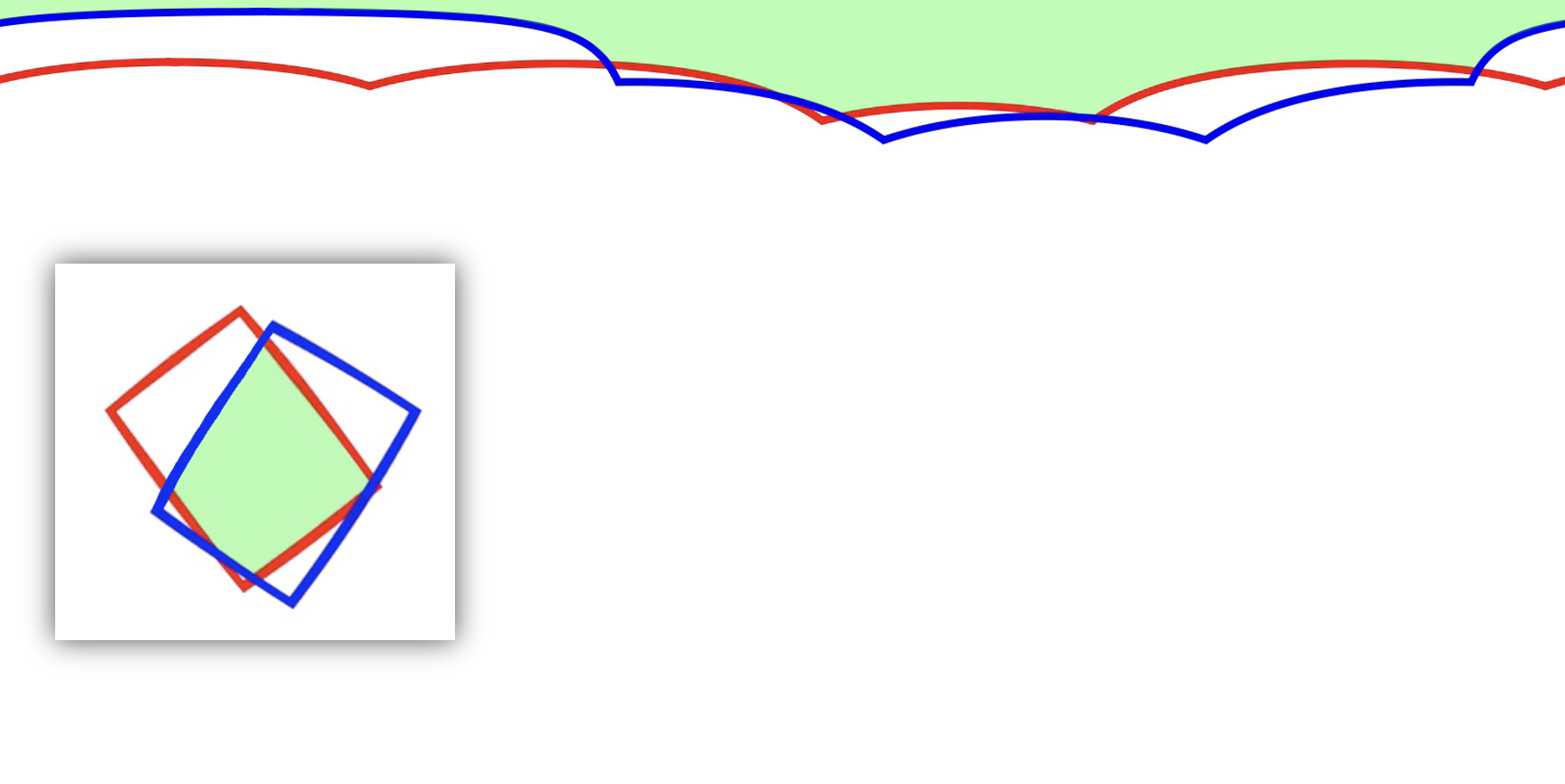}} \\
    \hline  
     Sph-IoU & 0.0 (-0.576) & 0.073 (-0.194) & 0.212 (-0.308) \\
     \hline  
     FoV-IoU & 0.589 (+0.013) & 0.259 (-0.012) & 0.538 (+0.018) \\
     \hline  
     Exact IoU & 0.576 & 0.267 & 0.52 \\
    \bottomrule
    \end{tabular}
    \label{tab:cases}
\end{table*}

\begin{table}[t]
    \centering
    \caption{Comparison of computational time. Avg time is the average time for single IoU computation. Frame per second (FPS) is the inference speed when using different types of IoU. }
    \begin{tabular}{c c c}
    \toprule
         & Avg. time (ms) & FPS  \\
         \midrule
        Exact IoU & 6.74 & 0.1  \\
        FoV-IoU & 0.14 & 7.1  \\
        Sph-IoU & 0.12 & 8.7  \\
        \bottomrule
    \end{tabular}
    \label{tab:computational_time}
\end{table}

We explained theoretically in the previous sections that FoV-IoU is more accurate than Sph-IoU as an approximation of spherical polygons (i.e., exact IoU). To further compare the computational accuracy of FoV-IoU and Sph-IoU experimentally, we compared FoV-IoU, Sph-IoU and exact IoU for six different arrangements of FoV-BBs in Table~\ref{tab:cases}. We observe that as the latitudes of BBs are higher, the less accurate the Sph-IoU becomes. In some cases, the error of Sph-IoU even reaches 0.5. Although both FoV-IoU and Sph-IoU have some deviations from the exact IoU, FoV-IoU is close to the exact value at different latitudes, and it is consistently more accurate than Sph-IoU.

As a computationally expensive part of the detection network, a time-consuming IoU computation can lead to an unacceptable training, inference and evaluation time. To test the feasibility of FoV-IoU, we also compared the computational time of single FoV-IoU, Sph-IoU, and exact IoU and the inference time using them. The results are shown in Table~\ref{tab:computational_time}. As expected, the computation of the exact IoU by the spherical polygon formula is quite time-consuming and leads to unacceptable training and inference time for the object detection task. Sph-IoU is the most efficient among them since it only requires the computation of the Euclidean distance, unlike FoV-IoU which requires the FoV distance computation. However, Sph-IoU is not an appropriate approximation of the exact IoU, which makes the detection results unreliable especially at high latitudes. On the other hand, FoV-IoU shows an efficiency practically comparable to Sph-IoU but has a more accurate approximation of the exact IoU. Considering the trade-off between accurate IoU calculation and computational efficiency, we use FoV-IoU to calculate the AP score in the following experiments.

\begin{figure}[t]
    \centering
    \includegraphics[width=1.0\linewidth]{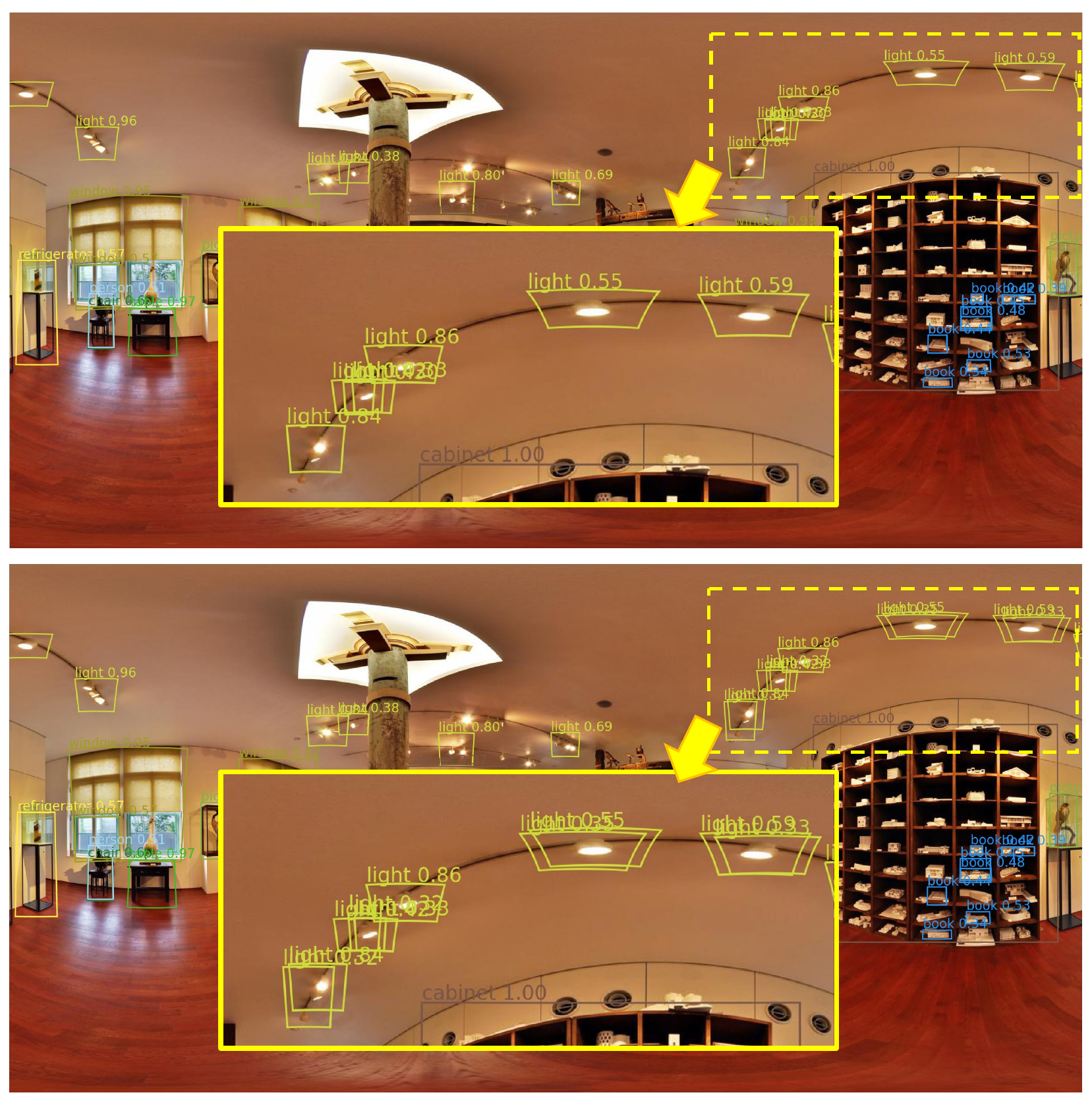}
    \caption{Non-maximum Suppression results by FoV-IoU (top) v.s. Sph-IoU (bottom).}
    \label{fig:nms}
\end{figure}

\subsubsection{FoV-IoU for the inference stage}
We illustrated some visualized examples in Fig.~\ref{fig:nms} to show the effect of FoV-IoU in the inference stage. The qualitative results show that NMS based on FoV-IoU can better filter the redundant predictions at high-latitude region. Since Sph-IoU underestimates the IoU value, some redundant predictions cannot be filtered out. On the other hand, FoV-IoU produce a more reasonable IoU value in NMS and redundant BBs are filtered out at different latitudes of the image. For a fair comparison in the training stage, we will uniformly use Sph-IoU to calculate NMS in the following experiments.

\subsubsection{FoV-GIoU Loss}

\begin{table}[t] 
    \centering
    \caption{Efficacy of FoV-GIoU with different object detectors. $\mathcal{L}$ with check-mark denotes training with FoV-GIoU loss and without it denotes with Sph-GIoU loss. Results are on both whole image and high-latitude (50\textdegree~to 90\textdegree) area. AP scores are calculated based on FoV-IoU.}
    \begin{tabular}{c|c|p{0.6cm} p{0.6cm} p{0.6cm}|p{0.6cm} p{0.6cm} p{0.6cm}}
    \toprule
    & & \multicolumn{3}{c|}{High latitude} & \multicolumn{3}{c}{Overall latitude} \\
    \midrule
    Detector & $\mathcal{L}$ & $AP$ & $AP_{50}$ & $AP_{75}$ & $AP$ & $AP_{50}$ & $AP_{75}$ \\
    \midrule
            Cascade & & 12.0 & 29.7 & 6.9 & 17.3 & 36.8 & 14.0 \\
        R-CNN & \Checkmark & \textbf{15.1} & \textbf{34.4} & \textbf{11.0} & \textbf{18.3} & \textbf{38.2} & \textbf{15.3} \\  
        \midrule
        Faster  & & 12.6 & 24.1 & 12.1 & 17.9 & \textbf{38.5} & 14.1  \\
        R-CNN & \Checkmark & \textbf{13.2} & \textbf{26.4} & \textbf{12.5} & \textbf{18.1} & \textbf{38.5} & \textbf{14.6}
        \\
         \midrule
        \multirow{ 2}{*}{FCOS}  & & 11.8 & 26.3 & \textbf{9.9} & 15.0 & 33.1 & \textbf{12.0} \\
          & \Checkmark & \textbf{13.7} & \textbf{30.7} & 8.0 & \textbf{15.4} & \textbf{34.1} & 11.8 \\
          \midrule
        \multirow{ 2}{*}{ATSS}  & & 11.6 & 28.9 & 7.9 & 17.3 & 34.6 & 15.4 \\
          & \Checkmark & \textbf{12.7} & \textbf{29.1} & \textbf{9.8} & \textbf{17.9} & \textbf{36.2} & \textbf{15.6} \\
        \midrule
        \multirow{ 2}{*}{YOLOv3} &  & 10.9 & 23.5 & 8.3  & 14.1 & 29.2 & \textbf{12.4} \\
         & \Checkmark & \textbf{12.2} & \textbf{25.5} & \textbf{10.9} & \textbf{14.3} & \textbf{29.6} & \textbf{12.4}  \\
    \bottomrule
    \end{tabular}
    \label{tab:exp_fovloss}
\end{table}

To demonstrate the advantage of FoV-GIoU loss over Sph-GIoU loss, we trained different object detectors, including Cascade R-CNN~\cite{cascade}, Faster R-CNN~\cite{fasterrcnn}, FCOS~\cite{fcos}, ATSS~\cite{atss}, and YOLOv3~\cite{yolo}, with FoV-GIoU loss and Sph-GIoU loss without changing other conditions (e.g., the architecture nor learning schedule). For $AP$ calculation, we used FoV-IoU and showed scores computed from objects from entire images and ones only from high-latitude areas (latitude from 50\textdegree~to 90\textdegree) to illustrate how FoV-IoU contributes to the robustness to projective distortions. 

The results are illustrated in Table~\ref{tab:exp_fovloss}. While our FoV-GIoU loss contributes to consistently improving baseline detectors, we observe that the advantage of FoV-GIoU loss over Sph-GIoU loss is more significant when detecting objects at high latitudes than detecting objects from an entire image. This is entirely reasonable because, as we have discussed in Section~\ref{sec:advantages}, Sph-IoU introduces underestimation bias due to the inaccurate approximation of the spherical polygons at higher latitudes, and our more accurate approximation could suppress this problem.

\subsection{Efficacy of 360Augmentation}
\begin{table}[!t] 
    \centering
    \footnotesize
    \caption{Comparison of 2D geometric transformation-based augmentation and 360Augmentaion. AP scores are calculated based on FoV-IoU.}
    \begin{tabular}{c p{0.6cm} p{0.6cm} p{0.6cm} p{0.6cm} p{0.6cm} p{0.6cm}}
    \toprule
    Aug options & $AP$ & $AP_{50}$ & $AP_{75}$ & $AP_{s}$ & $AP_{m}$ & $AP_{l}$ \\
    \midrule
        Vanilla & 17.9 & 38.5 & 14.1 & 3.7 & 13.5 & 27.6 \\
        2D Rotation  & 17.0 & 37.1 & 13.6 & 2.7 & 12.7 & 25.3 \\
        2D Translation & 16.9 & 37.3 & 13.1 & 2.6 & 12.8 & 25.5 \\
        360Augmentation & \textbf{18.9} & \textbf{41.1} & \textbf{14.7} & \textbf{3.6} & \textbf{15.2} & \textbf{27.8} \\
    \bottomrule
    \end{tabular}
    \label{tab:aug_compare}
\end{table}

\begin{table*}[h] 
    \centering
    \caption{Efficacy of FoV-GIoU with different object detectors. $\mathcal{L}$ with check-mark denotes training with FoV-GIoU loss and without it denotes with Sph-GIoU loss. Results are on both whole image and high-latitude (50\textdegree~to 90\textdegree) area. AP scores are calculated based on FoV-IoU.}
    \begin{tabular}{c|c|p{1cm}p{1cm}p{1cm}|p{1cm}p{1cm}p{1cm}}
    \toprule
    &  & \multicolumn{3}{c}{High-latitude} & \multicolumn{3}{|c}{Overall latitude} \\
    \midrule
    Detector & Ours & $AP$ & $AP_{50}$ & $AP_{75}$ & $AP$ & $AP_{50}$ & $AP_{75}$ \\
    \midrule
    
        Faster R-CNN & & 12.6 & 24.1 & 12.1 &  17.9 & 38.5 & 14.1 \\
        (ResNet-50) & \Checkmark & \textbf{14.1} & \textbf{31.8} & \textbf{12.2} & \textbf{19.1} & \textbf{40.8} & \textbf{15.7} \\  
        \midrule
        YOLOv3 & & 10.9 & 23.5 & 8.3 & 14.1 & 29.2 & 12.4  \\  
        (DarkNet-53) & \Checkmark & \textbf{13.8} & \textbf{30.3} & \textbf{10.1} & \textbf{16.2} & \textbf{33.8} & \textbf{14.3} \\  
        \midrule
    
        Cascade R-CNN & & 12.0 & 29.7 & 6.9 & 17.3 & 36.8 & 14.0  \\
        (ResNet-50) & \Checkmark & \textbf{13.2} & \textbf{30.5} & \textbf{8.6} & \textbf{18.4} & \textbf{38.7} & \textbf{14.9} \\  
        \midrule
        
        Cascade R-CNN & & 14.4 & 31.5 & 11.1 & 18.3 & 38.6 & 15.3  \\  
         (ResNet-101) & \Checkmark & \textbf{15.7} & \textbf{33.4} & \textbf{13.4} & \textbf{19.7} & \textbf{40.8} & \textbf{16.5} \\  
        \midrule

        FCOS & & 11.8 & 26.3 & 9.9 & 15.0 & 33.1 & 12.0  \\
        (ResNet-50) & \Checkmark & \textbf{13.6} & \textbf{32.9} & \textbf{10.6} & \textbf{16.9} & \textbf{37.2} & \textbf{13.2} \\  
        \midrule
        FCOS & & 10.9 & 26.2 & 9.0 & 15.6 & 33.3 & 12.7  \\  
        (ResNet-101) & \Checkmark & \textbf{13.9} & \textbf{28.9} & \textbf{12.6} & \textbf{17.1} & \textbf{36.8} & \textbf{14.1} \\  
        \midrule
        ATSS & & 11.6 & 28.9 & 7.9 & 17.3 & 34.6 & 15.4  \\
        (ResNet-50) & \Checkmark & \textbf{14.0} & \textbf{29.9} & \textbf{10.9} & \textbf{18.5} & \textbf{37.2} & \textbf{16.2} \\  
        \midrule
        ATSS & & 12.9 & 30.2 & 10.1 & 18.9 & 38.1 & 17.2  \\  
        (ResNet-101) & \Checkmark & \textbf{16.0} & \textbf{33.8} & \textbf{12.2} & \textbf{19.8} & \textbf{39.5} & \textbf{17.6} \\  
    \bottomrule
    \end{tabular}
    \label{tab:all}
\end{table*}

To verify the effectiveness of 360Augmentation, we compared Faster R-CNN models trained with 360Augmentation or 2D-based geometric transformation methods, including 2D rotation and 2D translation as comparisons. For all methods, we added $50\%$ of augmented images. The maximum angle for 2D rotation is 30\textdegree~and the maximum pixel's offset for 2D translation is 250. As shown in Table~\ref{tab:aug_compare}, the results of 2D-based geometric transformation methods are even lower than the vanilla method without any augmentation. After the 2D-based geometric transformation, the images no longer follow the spherical correspondence, resulting in inconsistent data between the training and testing stages. In contrast, 360Augmentation, as a spherical-based geometric transformation, retains the characteristics of 360-degree images while increasing the diversity of training data, resulting in a significant improvement over baseline. We should note that due to the characteristic of 360Augmentation, this improvement is observed not only at high latitudes but also from the entire coordinates.
\subsection{Different Choices of the Object Detector}
To demonstrate the performance of our complete package, we integrated both FoV-GIoU loss and 360Augmentation with several state-of-the-art object detectors, including two-stage detectors: Faster R-CNN~\cite{fasterrcnn}, Cascade R-CNN~\cite{cascade} and one-stage detectors: YOLOv3~\cite{yolo}, FCOS~\cite{fcos}, ATSS~\cite{atss}. $AP$ was calculated based on FoV-IoU for all original and integrated detectors. For a fair comparison, we trained and tested all the models with the default settings in mmdetection.

The results are shown in Table~\ref{tab:all}. We observe that the joint usage of proposed methods significantly improves all of the baselines by 0.9\% to 2.1\% overall $AP$ scores and 1.2\% to 3.1\% high-latitude $AP$ scores. Our method consistently boosts performance, especially in high-latitude areas. We analyzed that 360Augmentation improved both high-latitude and overall AP scores significantly since data augmentation is effective at all positions in the image. However, 360Augmentation alone cannot handle the high latitude distortions during training and test, so further improvement of accuracy can be expected by using the FoV-GIoU loss function at the same time.


\subsection{Comparison with methods specializing in 360° images}
    
\begin{table}[t] 
    \centering
    \footnotesize
    \caption{Results of comparison with 360\textdegree~object detection methods based on Faster R-CNN with MSCOCO pre-trained ResNet-50 backbone. AP scores are calculated based on FoV-IoU.}
    \begin{tabular}{c p{0.7cm} p{0.7cm} p{0.7cm} p{0.7cm} p{0.7cm} p{0.7cm}}
    \toprule
    Methods & $AP$ & $AP_{50}$ & $AP_{75}$ & $AP_{s}$ & $AP_{m}$ & $AP_{l}$ \\
    \midrule
        ERP  & 17.9 & 38.5 & 14.1 & 3.7 & 13.5 & 27.6 \\
        CubeMap & 14.5 & 35.9 & 8.9 & 3.4 & 13.5 & 23.6 \\
        S2CNN  & 10.9 & 24.5 & 8.3 & 0.2 & 5.3 & 20.4 \\
        SphereNet & 17.1 & 36.4 & 13.9 & 2.2 & 13.1 & 25.5 \\
        Ours & \textbf{19.1} & \textbf{40.8} & \textbf{15.7} & \textbf{4.7} & \textbf{14.9} & \textbf{27.2} \\
    \bottomrule
    \end{tabular}
    \label{tab:other360}
\end{table}

\begin{figure*}[t]
\begin{center}
\includegraphics[width=0.45\linewidth]{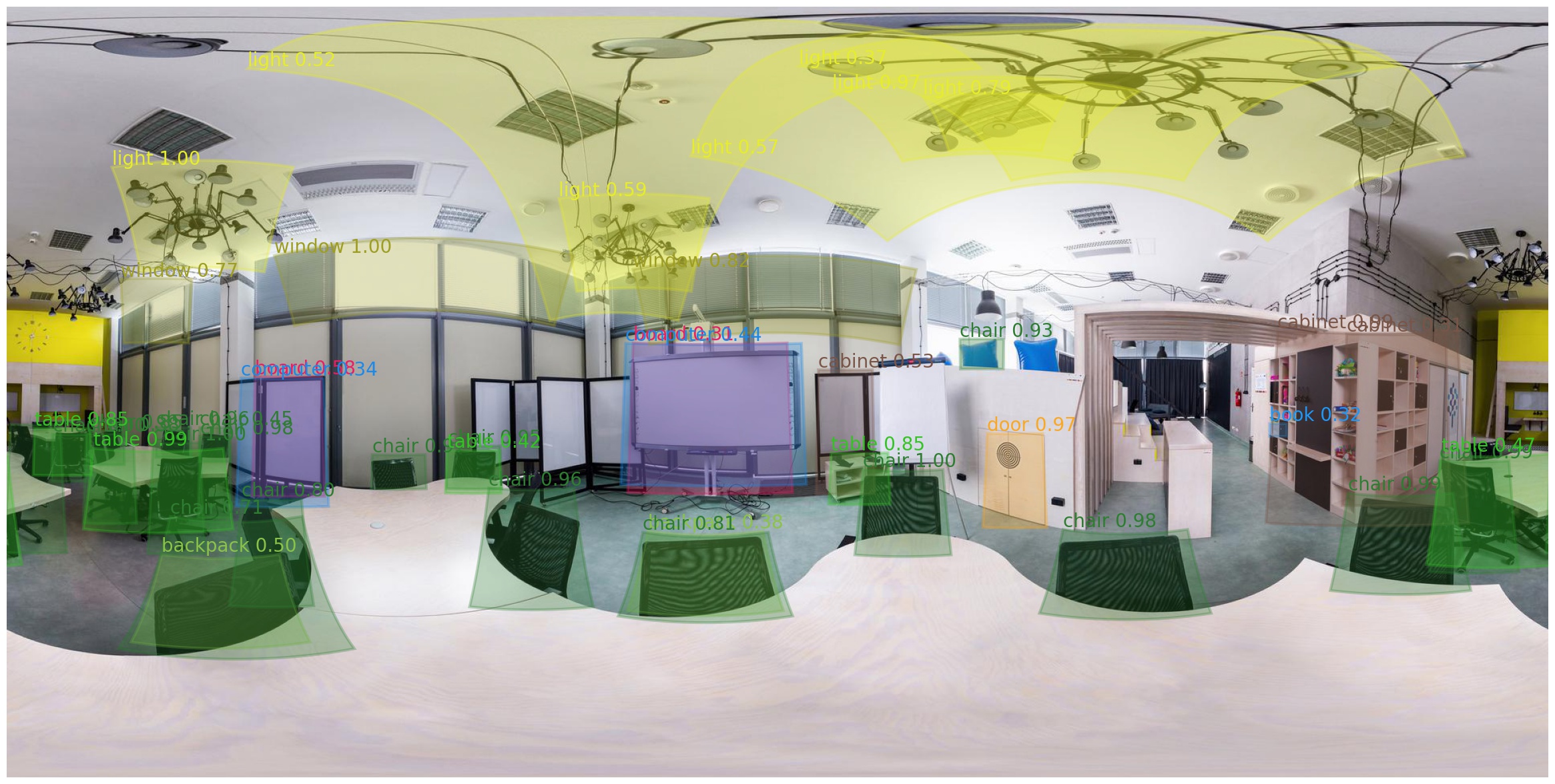}
\includegraphics[width=0.45\linewidth]{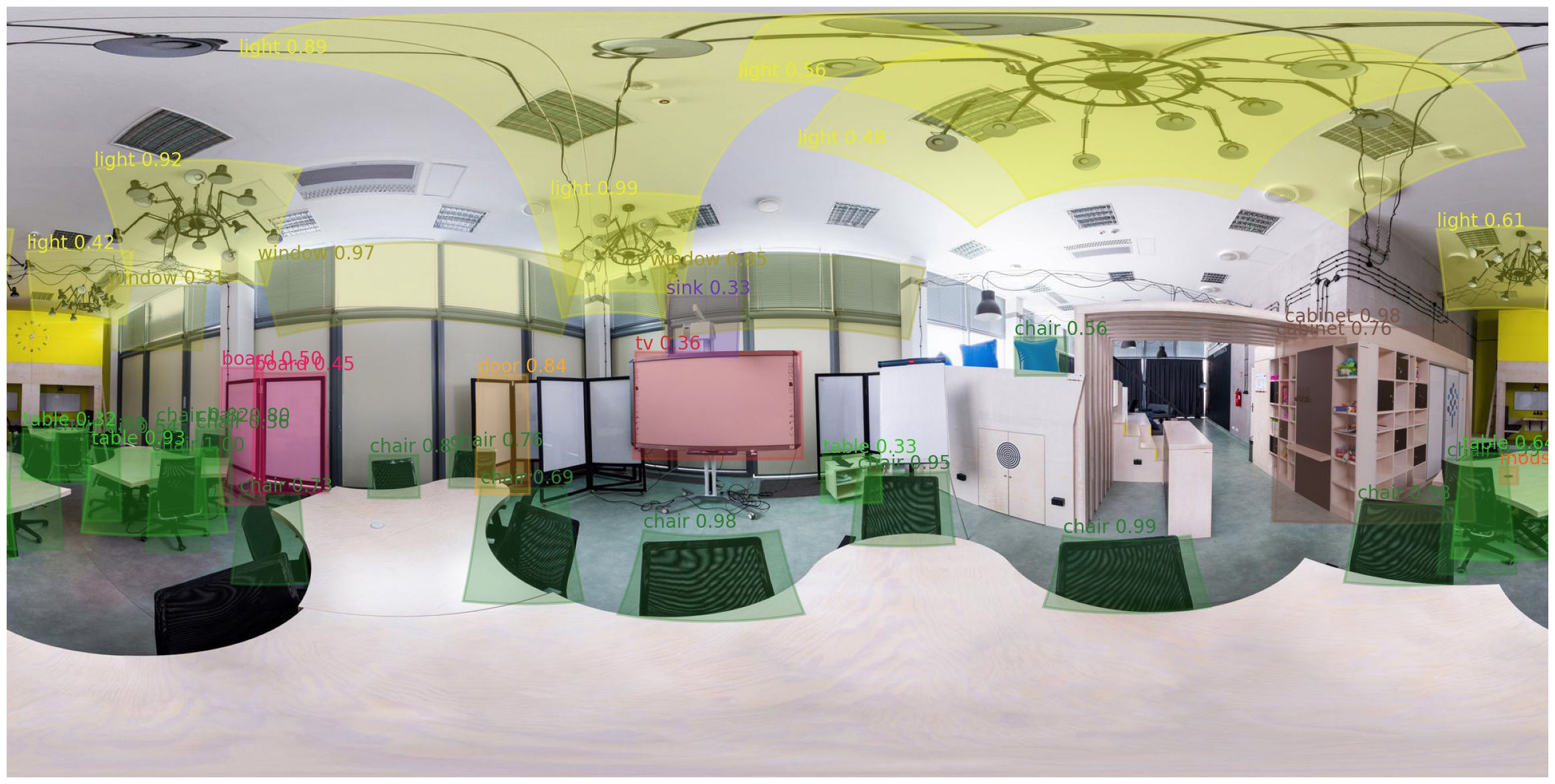}
\includegraphics[width=0.45\linewidth]{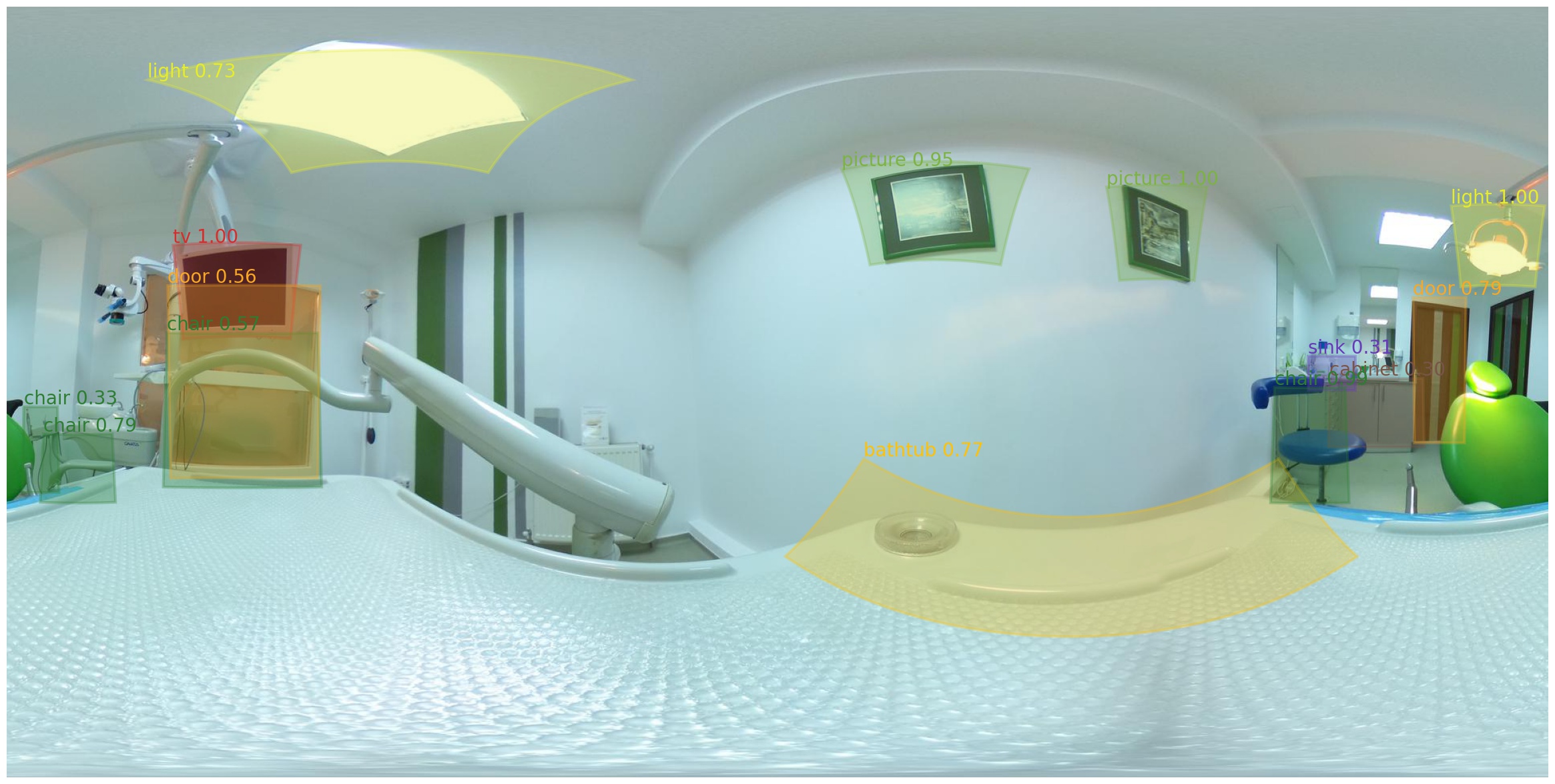}
\includegraphics[width=0.45\linewidth]{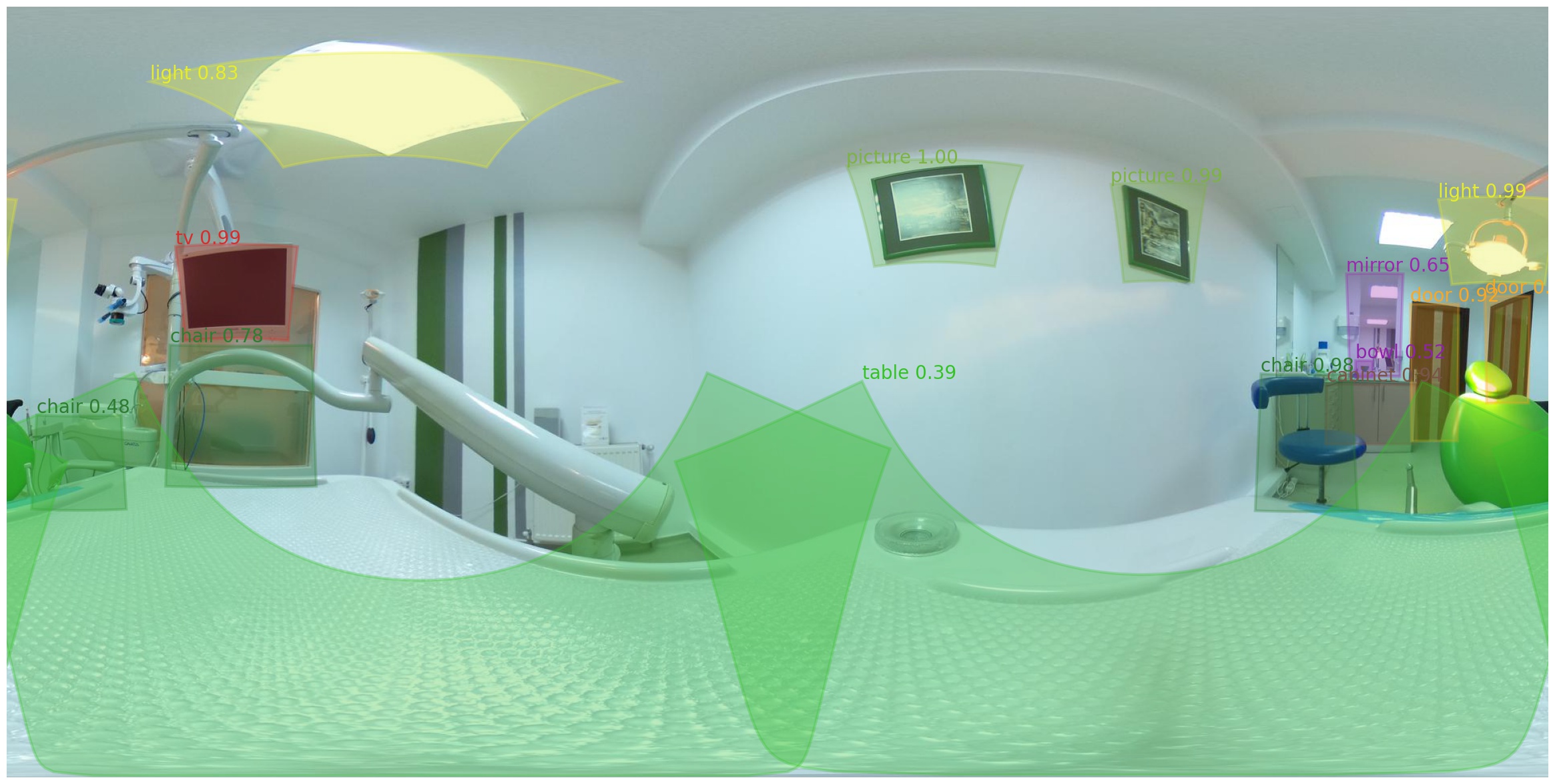}
\includegraphics[width=0.45\linewidth]{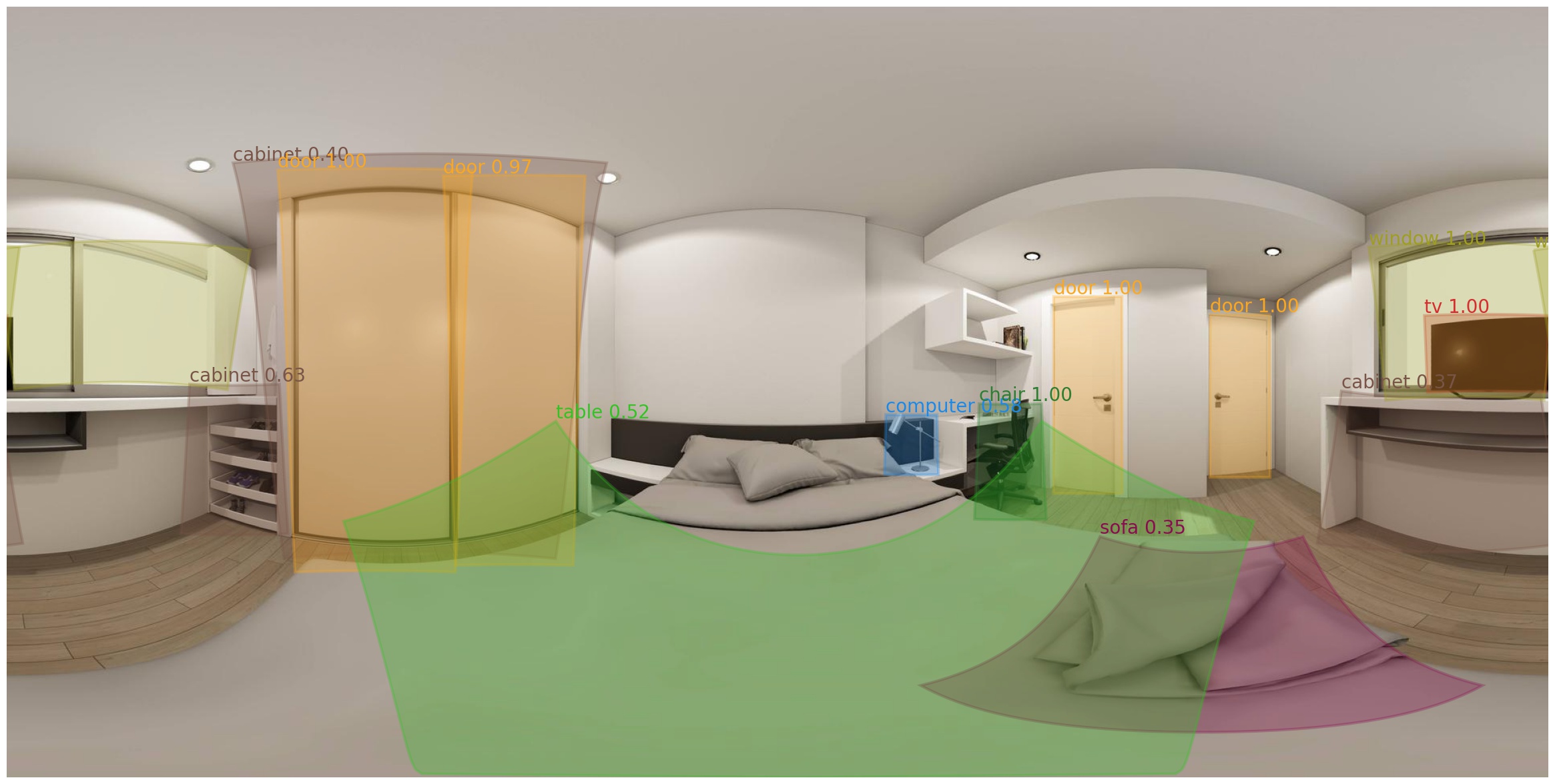}
\includegraphics[width=0.45\linewidth]{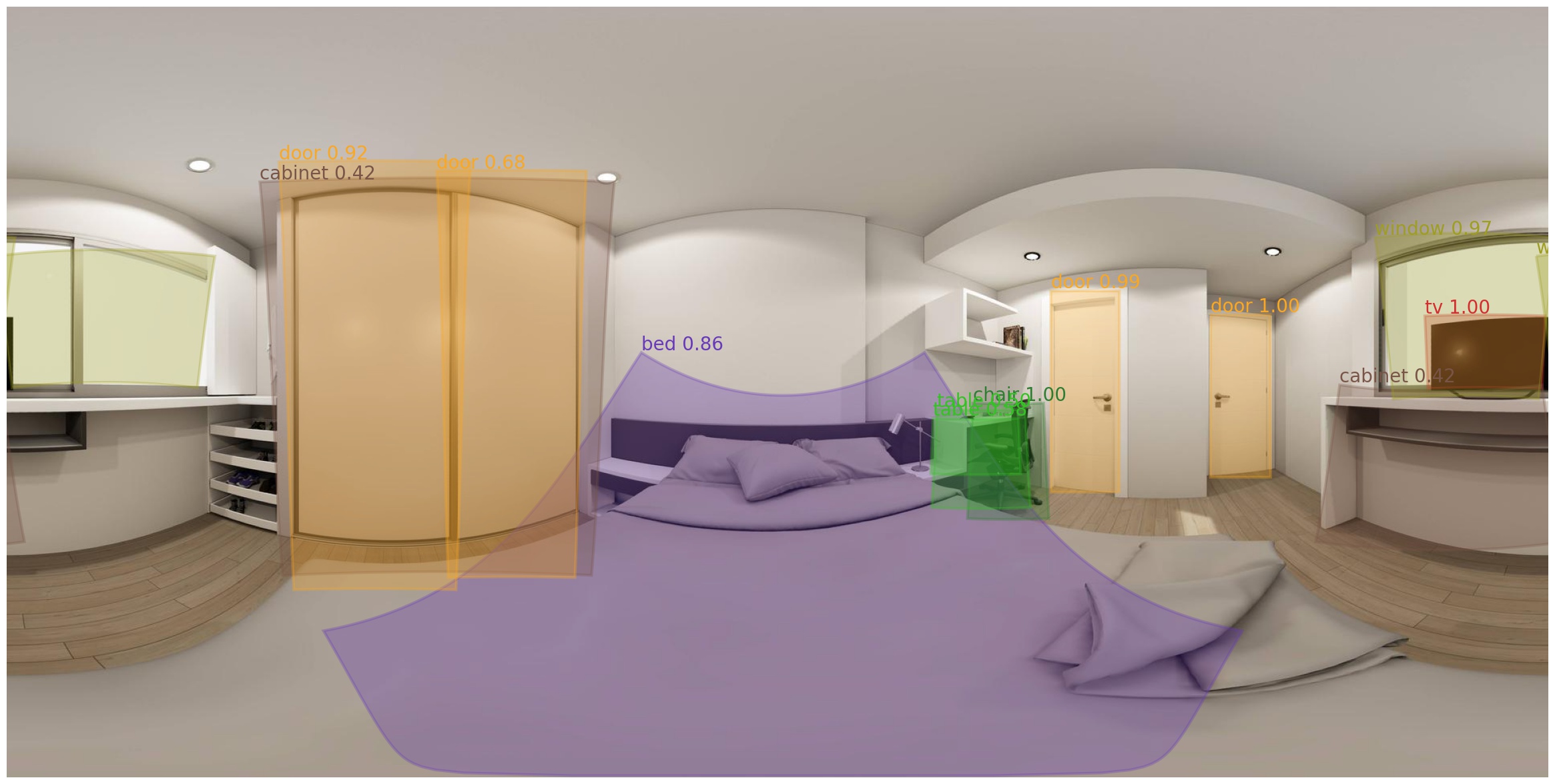}
\includegraphics[width=0.45\linewidth]{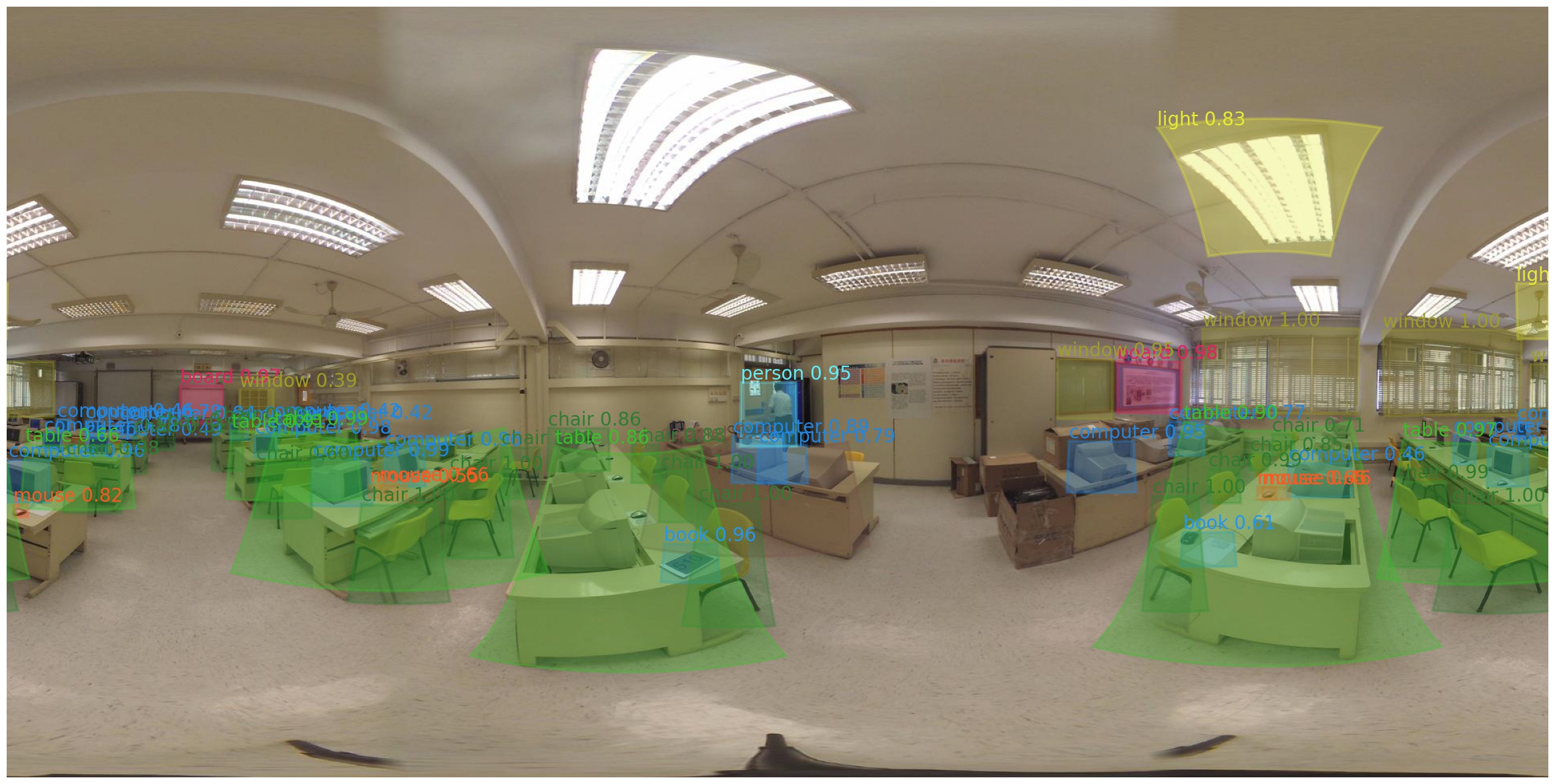}
\includegraphics[width=0.45\linewidth]{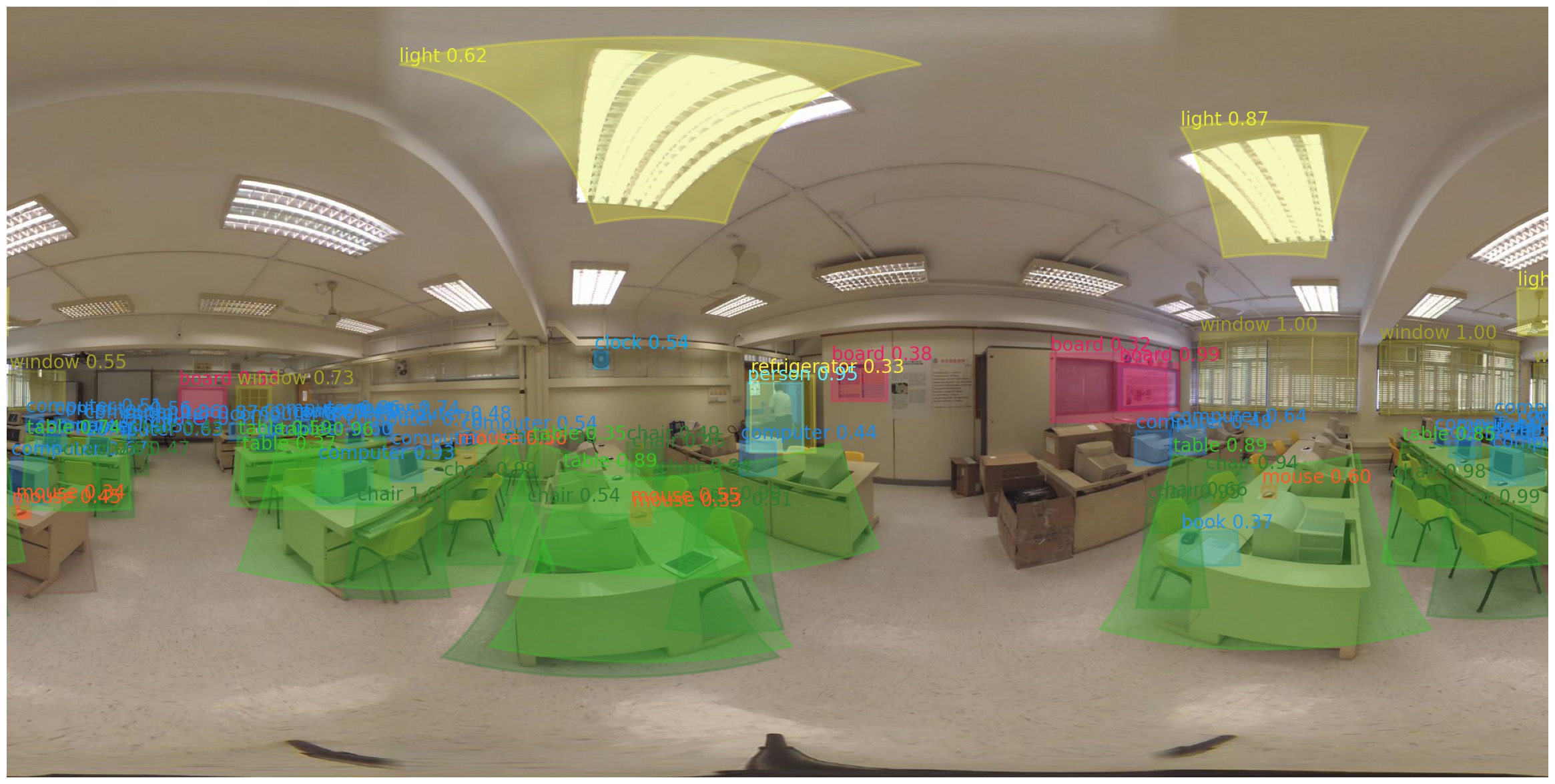}
\end{center}
  \caption{Visualization examples from Faster R-CNN trained on ERP directly (\textbf{left}) v.s. Faster R-CNN trained with FoV-GIoU loss and 360Aug (\textbf{right}).}
\label{fig:vis}
\end{figure*}

Finally, we compared our method (i.e., perspective object detectors trained with FoV-IoU loss and 360 Augmentation) against architectures originally designed for the 360\textdegree~image. For a fair comparison, we integrated them into the same backbone network architecture (i.e., Faster R-CNN with popular ResNet-50 backbone). Here we compared our method against S2CNN~\cite{cohen}, Faster R-CNN on the cube mapped images (CubeMap), SphereNet~\cite{coors2018spherenet} and Faster R-CNN on the ERP image without any modification (ERP). Here we briefly describe the implementation details of each algorithm.
\\
\noindent\textbf{ERP:} We trained a Faster R-CNN with on ERP images directly.
\\
\noindent\textbf{CubeMap:} Images and annotations in 360-indoor dataset were converted into CubeMap format as is same manner with~\cite{cubemap}, then we trained Faster R-CNN on converted images.
\\
\noindent\textbf{S2CNN~\cite{cohen}:} S2CNN was originally proposed for the 360\textdegree~image classification which is not possible to take high resolution images directly as input. Therefore, to validate the power of the spherical convolution ($S^2 conv$), we replaced the 3x3 convolutions of the last stage of ResNet by $S^2 conv$ and after each $S^2 conv$, the coordinate remapping was applied. Then, the output feature maps were passed to the detection head of Faster R-CNN 
for bounding box regression and classification. 
\\
\noindent\textbf{SphereNet~\cite{coors2018spherenet}:} 
SphereNet was originally implemented in SSD~\cite{ssd} with VGG-16~\cite{vgg} backbone. Since SphereNet only changed sampling locations of convolution and has the same input and output with normal convolution kernel, it can be directly integrated with other backbones. For fair comparison in our experiment, we integrated SphereNet in ResNet-50 backbone by replacing all 3x3 convolution kernels with SphereNet kernels, and trained SphereNet-integrated Faster R-CNN on ERP images.
\\

\noindent\textbf{Ours:} We integrated our proposed methods including FoV-IoU as loss function and 360Augmentation to Faster R-CNN and trained the model on ERP images.

The result is shown in Table~\ref{tab:other360}. Surprisingly, the Faster R-CNN trained on ERP images directly with ResNet-50 backbone outperformed other methods integrated with Faster R-CNN. We can see that CubeMap is worse than ERP possibly because CubeMap separated 360\textdegree~image in six perspective images, which broke the continuity in the ERP image. Another interesting observation is that the method using the spherical convolution (i.e., S2CNN, SphereNet) performed much worse than the vanilla Faster R-CNN. This is due to the fact that as the spherical CNN contributes to the rotation invariant tasks such as the image recognition but rather degrades the performance of rotation equivariant tasks such as object detection~\cite{chou2020indoor}.

In Fig.~\ref{fig:vis}, we show more visualized examples of predictions from original Faster R-CNN and Faster R-CNN trained with proposed FoV-GIoU loss and 360Aug. As shown in the figure on the left, the original detector is more likely to miss or misidentify objects that are at high-latitudes (e.g., tables, beds, lights). Qualitative results show that the proposed training strategy enhances the ability to recognize high-latitude objects correctly.


\section{Conclusion}
In this work, we proposed FoV-IoU which is an efficient and accurate way to compute IoU between FoV-BB. FoV-IoU provides a fairer evaluation score, better optimization loss, and effective NMS at inference for 360\textdegree~object detection. We also proposed 360Augmentation, a simple yet effective augmentation method specifically for 360\textdegree~object detection. Our methods can be easily integrated with different types of perspective object detectors. The experimental results show that our proposed methods consistently boost the performance of state-of-the-art object detectors in 360\textdegree~images. 

The limitation is, however, both FoV-IoU and Sph-IoU are still an approximation of the spherical polygons and inevitably misevaluate the exact intersection areas. Our future work is to improve the approximation accuracy for better object detection in 360\textdegree~images.

%

\bibliographystyle{IEEEtran}
\bibliography{ref}

\begin{thebibliography}{10}
\providecommand{\url}[1]{#1}
\csname url@samestyle\endcsname
\providecommand{\newblock}{\relax}
\providecommand{\bibinfo}[2]{#2}
\providecommand{\BIBentrySTDinterwordspacing}{\spaceskip=0pt\relax}
\providecommand{\BIBentryALTinterwordstretchfactor}{4}
\providecommand{\BIBentryALTinterwordspacing}{\spaceskip=\fontdimen2\font plus
\BIBentryALTinterwordstretchfactor\fontdimen3\font minus
  \fontdimen4\font\relax}
\providecommand{\BIBforeignlanguage}[2]{{%
\expandafter\ifx\csname l@#1\endcsname\relax
\typeout{** WARNING: IEEEtran.bst: No hyphenation pattern has been}%
\typeout{** loaded for the language `#1'. Using the pattern for}%
\typeout{** the default language instead.}%
\else
\language=\csname l@#1\endcsname
\fi
#2}}
\providecommand{\BIBdecl}{\relax}
\BIBdecl

\bibitem{icpr2018}
W.~Yang, Y.~Qian, J.-K. Kämäräinen, F.~Cricri, and L.~Fan, ``Object
  detection in equirectangular panorama,'' in \emph{2018 24th International
  Conference on Pattern Recognition (ICPR)}, 2018, pp. 2190--2195.

\bibitem{su2017watchable}
Y.-C. Su and K.~Grauman, ``Making 360° video watchable in 2d: Learning
  videography for click free viewing,'' in \emph{2017 IEEE Conference on
  Computer Vision and Pattern Recognition (CVPR)}, 2017, pp. 1368--1376.

\bibitem{yu2018deep}
Y.~Yu, S.~Lee, J.~Na, J.~Kang, and G.~Kim, ``A deep ranking model for
  spatio-temporal highlight detection from a 360 video,'' 2018.

\bibitem{icvrv2017}
Y.~Zhang, X.~Xiao, and X.~Yang, ``Real-time object detection for 360-degree
  panoramic image using cnn,'' in \emph{2017 International Conference on
  Virtual Reality and Visualization (ICVRV)}, 2017, pp. 18--23.

\bibitem{su2017}
Y.-C. Su and K.~Grauman, ``Learning spherical convolution for fast features
  from 360° imagery,'' in \emph{NeurIPS}, 2017.

\bibitem{coors2018spherenet}
B.~e. Coors, A.~P. Condurache, and A.~Geiger1, ``Spherenet: Learning spherical
  representations for detection and classification in omnidirectional images,''
  in \emph{ECCV}, 2018.

\bibitem{iouloss}
\BIBentryALTinterwordspacing
D.~Zhou, J.~Fang, X.~Song, C.~Guan, J.~Yin, Y.~Dai, and R.~Yang, ``Iou loss for
  2d/3d object detection,'' \emph{CoRR}, vol. abs/1908.03851, 2019. [Online].
  Available: \url{http://arxiv.org/abs/1908.03851}
\BIBentrySTDinterwordspacing

\bibitem{giou}
H.~Rezatofighi, N.~Tsoi, J.~Gwak, A.~Sadeghian, I.~Reid, and S.~Savarese,
  ``Generalized intersection over union: A metric and a loss for bounding box
  regression,'' in \emph{2019 IEEE/CVF Conference on Computer Vision and
  Pattern Recognition (CVPR)}, 2019, pp. 658--666.

\bibitem{zhao2020criteria}
P.~Zhao, A.~You, Y.~Zhang, J.~Liu, K.~Bian, and Y.~Tong, ``Spherical criteria
  for fast and accurate 360° object detection,'' in \emph{AAAI}, 2020.

\bibitem{chou2020indoor}
S.-H. Chou, C.~Sun, W.-Y. Chang, W.-T. Hsu, M.~Sun, and J.~Fu, ``360-indoor:
  Towards learning real-world objects in 360◦ indoor equirectangular
  images,'' in \emph{WACV}, 2020.

\bibitem{sphpoly}
I.~Todhunter, Ed., \emph{Spherical Trigonometry: For the Use of Colleges and
  Schools}.\hskip 1em plus 0.5em minus 0.4em\relax MACMILLAN AND CO., 1886.

\bibitem{hav}
C.~C. Robusto, ``The cosine-haversine formula,'' \emph{The American
  Mathematical Monthly}, vol.~64, no.~1, pp. 38--40, 1957.

\bibitem{coco}
T.-Y. Lin, M.~Maire, S.~Belongie, L.~Bourdev, R.~Girshick, J.~Hays, P.~Perona,
  D.~Ramanan, C.~L. Zitnick, and P.~Dollár, ``Microsoft coco: Common objects
  in context,'' in \emph{ECCV}, 2014.

\bibitem{chen2019mmdetection}
K.~Chen, J.~Wang, J.~Pang, Y.~Cao, Y.~Xiong, X.~Li, S.~Sun, W.~Feng, Z.~Liu,
  J.~Xu, Z.~Zhang, D.~Cheng, C.~Zhu, T.~Cheng, Q.~Zhao, B.~Li, X.~Lu, R.~Zhu,
  Y.~Wu, J.~Dai, J.~Wang, J.~Shi, W.~Ouyang, C.~C. Loy, and D.~Lin,
  ``Mmdetection: Open mmlab detection toolbox and benchmark,'' 2019.

\bibitem{ultralytics}
\BIBentryALTinterwordspacing
G.~Jocher, guigarfr, perry0418, Ttayu, J.~Veitch-Michaelis, G.~Bianconi,
  F.~Baltacı, D.~Suess, WannaSeaU, and IlyaOvodov, ``{ultralytics/yolov3:
  Rectangular Inference, Conv2d + Batchnorm2d Layer Fusion},'' Apr. 2019.
  [Online]. Available: \url{https://doi.org/10.5281/zenodo.2672652}
\BIBentrySTDinterwordspacing

\bibitem{cascade}
Z.~Cai and N.~Vasconcelos, ``Cascade r-cnn: Delving into high quality object
  detection,'' in \emph{Proceedings of the IEEE Conference on Computer Vision
  and Pattern Recognition (CVPR)}, June 2018.

\bibitem{fasterrcnn}
S.~Ren, K.~He, R.~Girshick, and J.~Sun, ``Faster r-cnn: Towards real-time
  object detection with region proposal networks,'' in \emph{NeurIPS}, 2015.

\bibitem{fcos}
Z.~Tian, C.~Shen, H.~Chen, and T.~He, ``Fcos: Fully convolutional one-stage
  object detection,'' in \emph{Proceedings of the IEEE/CVF International
  Conference on Computer Vision (ICCV)}, October 2019.

\bibitem{atss}
S.~Zhang, C.~Chi, Y.~Yao, Z.~Lei, and S.~Z. Li, ``Bridging the gap between
  anchor-based and anchor-free detection via adaptive training sample
  selection,'' in \emph{Proceedings of the IEEE/CVF Conference on Computer
  Vision and Pattern Recognition (CVPR)}, June 2020.

\bibitem{yolo}
J.~Redmon, S.~Divvala, R.~Girshick, and A.~Farhadi, ``You only look once:
  Unified, real-time object detection,'' in \emph{CVPR}, 2016.

\bibitem{cohen}
T.~S. Cohen, M.~Geiger, J.~Köhler, and M.~Welling, ``Spherical cnns,'' in
  \emph{ICLR}, 2018.

\bibitem{cubemap}
N.~Greene, ``Environment mapping and other applications of world projections,''
  \emph{IEEE Computer Graphics and Applications}, vol.~6, no.~11, pp. 21--29,
  1986.

\bibitem{ssd}
\BIBentryALTinterwordspacing
W.~Liu, D.~Anguelov, D.~Erhan, C.~Szegedy, S.~E. Reed, C.~Fu, and A.~C. Berg,
  ``{SSD:} single shot multibox detector,'' \emph{CoRR}, vol. abs/1512.02325,
  2015. [Online]. Available: \url{http://arxiv.org/abs/1512.02325}
\BIBentrySTDinterwordspacing

\bibitem{vgg}
\BIBentryALTinterwordspacing
K.~Simonyan and A.~Zisserman, ``Very deep convolutional networks for
  large-scale image recognition,'' in \emph{3rd International Conference on
  Learning Representations, {ICLR} 2015, San Diego, CA, USA, May 7-9, 2015,
  Conference Track Proceedings}, Y.~Bengio and Y.~LeCun, Eds., 2015. [Online].
  Available: \url{http://arxiv.org/abs/1409.1556}
\BIBentrySTDinterwordspacing

\end{thebibliography}

\vfill

\end{document}